%% file: 00-main.tex
\documentclass{article}

\PassOptionsToPackage{numbers, square, compress}{natbib}

\usepackage[final]{neurips_data_2024}

\usepackage[utf8]{inputenc} 
\usepackage[T1]{fontenc}    
\usepackage{hyperref}       
\usepackage{url}            
\usepackage{booktabs}       
\usepackage{amsfonts}       
\usepackage{nicefrac}       
\usepackage{microtype}      
\usepackage{xcolor}         
\usepackage{graphicx}
\usepackage{caption}
\usepackage{comment}
\usepackage{tabularx}
\usepackage{float}
\usepackage{xspace}
\usepackage{wrapfig}
\usepackage{tikz}
\usepackage{bm}
\usepackage{multirow}
\usepackage{titletoc}
\usepackage[group-separator={,}]{siunitx}
\usepackage{wrapfig}
\usepackage{subcaption}
\usepackage{outlines}
\usepackage{makecell}
\usepackage{booktabs}
\usepackage[export]{adjustbox}
\usepackage{csvsimple}
\usepackage{longtable}
\usepackage{layouts}
\usepackage{amssymb}
\usepackage{pifont}

\usetikzlibrary{positioning}
\usetikzlibrary{calc}
\usetikzlibrary{matrix}

\newcommand{\cmark}{\ding{51}}%
\newcommand{\xmark}{\ding{55}}

 \newcommand{\midsepdefault}{\aboverulesep = 0.605mm \belowrulesep = 0.984mm}

\usepackage{xspace}
\newcommand*{\eg}{e.g.\@\xspace}
\newcommand*{\ie}{i.e.\@\xspace}
\newcommand*{\cf}{c.f.\@\xspace}
\newcommand*{\pp}{p.p.\@\xspace}

\newcommand{\ignore}[1]{}

\newcommand{\dalle}{DALL·E 3\@\xspace}
\newcommand{\midjourney}{Midjourney~v6\@\xspace}
\newcommand{\imagenA}{Imagen-A\@\xspace}
\newcommand{\imagenB}{Imagen-B\@\xspace}
\newcommand{\imagenC}{Imagen-C\@\xspace}
\newcommand{\imagenD}{Imagen-D\@\xspace}
\newcommand{\imagen}{Imagen\@\xspace}
\newcommand{\geckonum}{\textsc{GeckoNum}\@\xspace}

\newcommand{\museA}{Muse-A\@\xspace}
\newcommand{\museB}{Muse-B\@\xspace}

\newcommand{\sdone}{SD1.5\@\xspace}

\newcommand{\sdthree}{SD3\@\xspace}
\newcommand{\sdxl}{SDXL\@\xspace}

\newcommand{\paligemma}{PaLIGemma\@\xspace}

\usepackage{array}
    \newcolumntype{C}{>{$}c<{$}}
    \newcolumntype{L}{>{$}l<{$}}
    \newcolumntype{R}{>{$}r<{$}}

\title{Evaluating Numerical Reasoning in Text-to-Image Models}

\author{%
  Ivana Kaji{\'c} \\
  Google DeepMind \\
   \And
   Olivia Wiles \\
  Google DeepMind \\
  \And
   Isabela Albuquerque\\
  Google DeepMind \\
   \AND
   Matthias Bauer \\
   Google DeepMind \\
   \And
   Su Wang \\
   Google DeepMind \\
    \And
    Jordi Pont-Tuset \\
    Google DeepMind \\
    \And
    Aida Nematzadeh \\
    Google DeepMind 
}

\begin{document}

\maketitle

\begin{abstract}
  Text-to-image generative models are capable of producing high-quality images that often faithfully depict concepts described using natural language. In this work, we comprehensively evaluate a range of text-to-image models on numerical reasoning tasks of varying difficulty, and show that even the most advanced models have only rudimentary numerical skills. Specifically, their ability to correctly generate an exact number of objects in an image is limited to small numbers, it is highly dependent on the context the number term appears in, and it deteriorates quickly with each successive number. We also demonstrate that models have poor understanding of linguistic quantifiers (such as “a few” or “as many as”), the concept of zero, and struggle with more advanced concepts such as partial quantities and fractional representations. We bundle prompts, generated images and human annotations into \geckonum, a novel benchmark for evaluation of numerical reasoning.
\end{abstract}

\input{01-intro}
\input{02-related-work}

\input{03-the-prompts}
\input{05-human-annotation}
\input{06-results}
\input{07-beyond-benchmark}
\input{08-discussion}

\bibliographystyle{abbrvnat}
\bibliography{00-main}
\clearpage
\input{0x-checklist}

\clearpage
\startcontents[appendices]
\appendix
\input{app_prompts}
\input{app_results}
\input{app_human_annotation}
\input{app_vqa}

\clearpage


\end{document}

%% file: 01-intro.tex
\section{Introduction}

Recent generative text-to-image models can produce images of impressive quality in a variety of styles and following the text descriptions provided by users~\cite{pmlr-v202-chang23b,Rombach2022CVPR,vasconcelos2024greedy}. However, they may still fail to accurately generate images where the given descriptions contain numbers and quantities (such as ``7 pistachios'', as shown in Figure~\ref{fig:results-task1-examples})~\cite{paiss2023teaching,rane2024can}. 
While recent work has focused on designing metrics, benchmarks and methods for evaluating specific capabilities of text-to-image generative models, such as alignment~\cite{wiles2024revisiting,hu2023tifa,cho2023davidsonian}, compositionality~\cite{huang2024t2i}, or spatial reasoning~\cite{gokhale2022benchmarking}, there is no comprehensive benchmark for evaluation of numerical reasoning.

We address this gap by proposing \geckonum, a comprehensive and controlled benchmark of text prompts aimed at evaluating different aspects of numerical reasoning in text-to-image models.
We formalize evaluation as three tasks: exact number generation, approximate number generation, and reasoning about partial quantities.
For each task, we design various template types to control for different variables such as sentence structure, the context in which the number words occur in, and the number of attributes/entities in a prompt. 
Table~\ref{tab:prompt_types} shows examples of numerical tasks and the associated prompt types.

Using \geckonum, we evaluate twelve models chosen from five different model families (\dalle, Midjourney, Imagen, Muse and Stable Diffusion): we generate images for these models and collect human annotations to measure whether the images correctly match the prompts with respect to numerical reasoning. Our benchmark consists of \num{1386} text prompts, \num{52721} generated images, and a total of \num{479570} human annotations that we release.\footnote{ \url{https://github.com/google-deepmind/geckonum_benchmark_t2i}} 

Overall, our results demonstrate that the recent generative text-to-image models have rudimentary numerical reasoning skills, and are most accurate when tested for generation of small exact quantities.
We highlight the utility of \geckonum as an evaluation benchmark: it can discriminate between models, even the powerful ones, such as Imagen-D~\cite{vasconcelos2024greedy} and \dalle~\cite{betker2023improving}, that are similar in terms of high image quality. Finally, we demonstrate that our benchmark can be used to drive progress in related research areas, such as the development of automatic evaluation metrics and evaluation and improvement of pretrained vision--language models on counting.

%% file: 02-related-work.tex
\section{Related Work}
\begin{figure}[t]
    \centering
    \includegraphics[width=\linewidth]{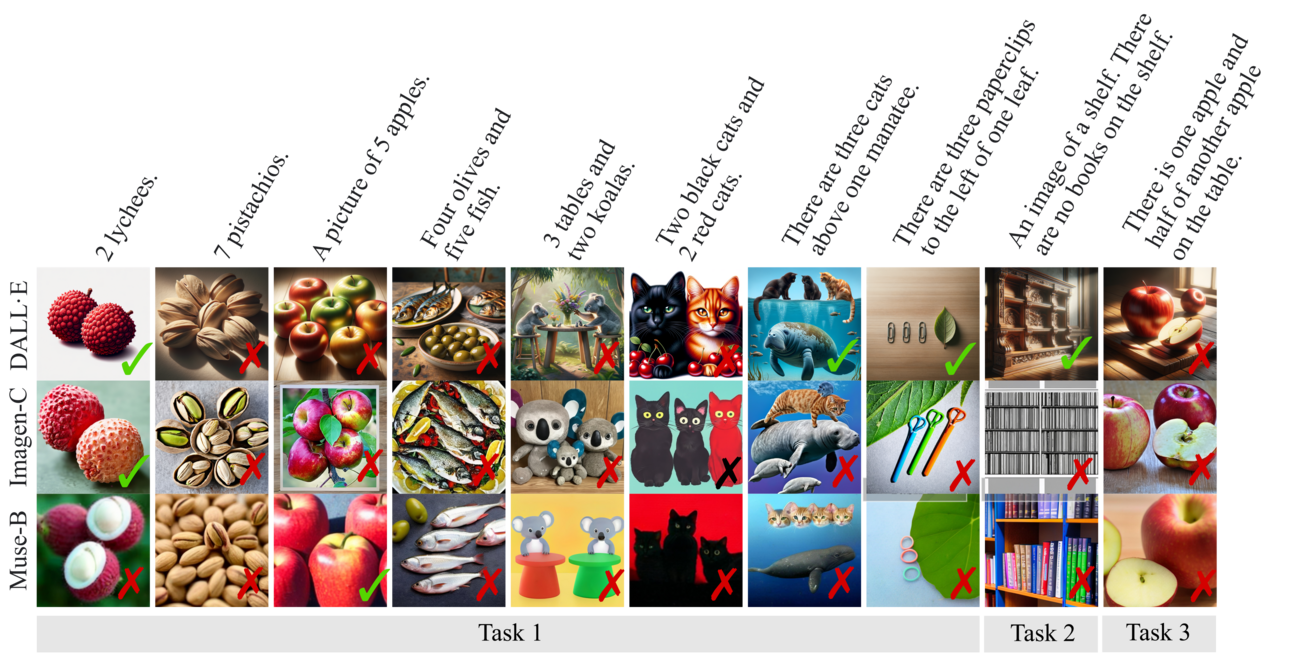}
    \caption{Examples of images generated by selected models: \dalle, \imagenC and \museB. Correctly generated images are marked with a check mark ``\cmark'', and incorrect with a cross mark ``\xmark''. } 
    \label{fig:results-task1-examples}
\end{figure}

{\bf Text-to-image benchmarking.}
Benchmarks used for evaluation of text-to-image models typically consist of a set of text prompts that target a specific capability. 

Some benchmarks, such as \mbox{\dalle} Eval \citep{betker2023improving} are more general and aim to capture real world use cases, while others contain a variety of challenges, such as prompts in DrawBench \citep{saharia2022photorealistic}, PartiPrompts \citep{yu2022scaling}, and HEIM \citep{lee2024holistic}.
Others are developed to interrogate models on more specific capabilities such as alignment (e.g.~TIFA \citep{hu2023tifa}, DSG \citep{cho2023davidsonian} and Gecko \citep{wiles2024revisiting}) or compositionality (e.g. T2I-CompBench \citep{huang2024t2i}).

Often, prompts in such benchmarks are curated from the data harvested from the Web, and they may include numbers or other numerical concepts relevant to evaluation of numerical reasoning.
However, the number of such prompts in existing benchmarks is often limited, and the  complexity of prompts may not be appropriate to evaluate models on numerical reasoning as a correct machine interpretation of such prompts often requires a combination of several different reasoning skills.
For example, the understanding of the number ``2'' in a simple prompt such as \emph{``Two zebras in Cape Town''} (from CountBench~\cite{paiss2023teaching}) also requires an interpretation of a geographical location.
For the correct interpretation of \emph{``Two dogs in a grassy field with one dog holding an orange disc.''} (from MS COCO~\cite{chen2015microsoft}) the model would need to correctly generate several objects (\ie \emph{``dog'', ``disc''}), relations between them (\emph{``holding''}), and correctly bind attributes (\emph{``orange''}) to objects. 
While the complexity of such prompts may be representative of the complexity observed in  natural language, it may hinder our ability to accurately evaluate numerical reasoning in text-to-image models.

The prompt set in \geckonum is vastly more comprehensive and it systematically covers various dimensions of evaluation that affect numerical reasoning, such as different number ranges, types of nouns, different ways of representing numbers, approximations based on linguistic quantifiers, and reasoning about partial quantities that are missing in other datasets. We show that to evaluate a specific capability thoroughly requires an extensive, comprehensive dataset. Most similar to our work is \cite{rane2024can}, which investigates number generation for text-to-image models. While~\cite{vasconcelos2024greedy} show that number generation improves with larger model sizes, their set of prompts is small (N=59) and focuses only on the simplest prompt structure.
We go beyond this by evaluating a comprehensive set of model families, considering other properties of numerical reasoning such as estimation and conceptual quantitative reasoning, and a more comprehensive breakdown of performance based on how commonplace an entity is.

{\bf Measuring counting in image-to-text models.}
While numerical reasoning has no standardized benchmark for text-to-image generation, there are some widely accepted benchmarks within the visual question answering domain. For example, TallyQA~\cite{acharya2019tallyqa} and CountBench~\cite{paiss2023teaching} both measure the counting ability of question-answering models.
CountBench is a small dataset of only 540 images whereas TallyQA contains approximately 20K images for evaluation.
However, despite TallyQA's size, the distribution of numbers is highly skewed towards small numbers (\eg 1, 2); and the quality of the images and associated labels is mixed.

%% file: 03-the-prompts.tex
\section{Tasks to Examine Numerical Reasoning}
\label{sec:curating_prompts}
\begin{table}[t]
\small
    \centering
    \caption{Twelve prompt types, example prompts and templates to probe different aspects of numerical reasoning in text-to-image models.}
    \footnotesize
    \include{./tables/prompt_examples}

    \label{tab:prompt_types}
\end{table}
\midsepdefault

We evaluate text-to-image models on different aspects of numerical reasoning formulated as three numerical tasks and a set of text prompt types for each task, spanning different levels of difficulty. 
Our working definition of numerical reasoning draws inspiration from literature in numerical cognition, and concerns both the ability to reason in abstract terms and the ability to manipulate such abstractions across different contexts.
Specifically, one fundamental aspect of abstract reasoning is the ability to form a representation of a set size independently of the identity of objects, known as \emph{The abstraction principle}~\cite{gelman1986child}. For example, understanding that ``two'' refers to the same quantity in ``Two apples'' as it does in ``Two letters'' even though apples and letters are different objects.
We first define the tasks and describe what aspect of numerical reasoning they intend to evaluate, followed by description of prompts that aim to require those aspects of reasoning. These prompts are generated using intentionally designed 12 templates where we sweep over combinations of numbers and selected word nouns, resulting in \num{1386} prompts in total.
The 12 prompt types, including prompts and example templates used to generate the prompts are shown in Table~\ref{tab:prompt_types}, with further details available in Table~\ref{tab:app_dist_prompts}.

\subsection{Task 1: Exact Number Generation} 
Task 1 examines a model's ability to correctly generate an exact number of objects.  We define number generation as the correct visual depiction of an entity specified in a prompt and its associated quantity (\eg,``2 red cats.'' or ``One mushroom and 3 koalas.'').

To probe how well models capture {\em The abstraction principle}, we vary the context of numerical terms appear in. 
Then, we investigate whether and how image generation accuracy changes depending on the prompt structure (\eg, attributes, and compositionality) enumerated below to give seven different prompt types listed together with examples in rows ``Exact'' in Table~\ref{tab:prompt_types}.
{\bf Prompt structure:} The simplest setting consists of phrases with an object and a number (\emph{numeric-simple}). 
We explore different prompt structure of sentences (\emph{numeric-sentence}), and also consider prompts which contain two or three number-noun combinations (\emph{2-additive} and \emph{3-additive}). We then include color adjectives for both simple prompts with one entity (\emph{attribute-color)}, and two entities with associated colors (\emph{2-additive-color}). Finally, we combine number terms with {spatial relationships} such as ``There are three cats above one manatee.'' (\emph{attribute-spatial}).

{\bf Exact number generation:} For the simplest setting (\emph{numeric-simple}), we additionally examine the role of three other factors: (i) \emph{Number magnitude} for which we generate text prompts with numbers ranging from 1 to 10. (ii) \emph{Number representation} where we consider prompts that represent digits with both Arabic numerals (\eg, 1, 2, 3) or words (\eg, ``one'', ``two'', ``three''). (iii) \emph{Noun frequency} in English for which we select nouns (\ie, entities) to cover both frequent and rare words.\footnote{We use the wordfreq Python library~\cite{robyn_speer_2022_7199437} to determine word frequency. Our vocabulary in this task consists of approx. $40$ words from four common categories: food, animals, nature, and objects. See Appendix~\ref{app:prompts} for more details.}

\subsection{Task 2: Approximate Number Generation and Zero}
In Task 2, we evaluate models on their ability to correctly depict entities with quantities expressed in \emph{approximate} terms by means of linguistic quantifiers (\eg, ``many'', ``a few'', or ``more'').
Such quantity terms are ubiquitous in ordinary language use and may denote a range of numbers thus carrying a more fuzzy interpretation of quantities. 
However, people tend to be relatively consistent when using such terms, as shown in Figure~\ref{fig:app_task2_counts_humans}.
We expect that models that correctly depict approximate quantities align with human perception of these quantities.
We also examine how well models interpret the concept of zero, which we evaluate separately as it represents a challenging milestone in number learning in children~\cite{wellman1986thinking}.

\noindent\textbf{Prompt types.} We design two prompt templates for this purpose: \emph{approx-1-entity} and \emph{approx-2-entity}, as shown in Table~\ref{tab:prompt_types}. The former tests for approximate generation of one entity in the prompt (\eg, a few \emph{candles}); the latter includes two entities with quantities expressed in relation to each other (\eg, more \emph{books} than \emph{cats}), as existing studies indicate poor performance on this prompt type~\cite{leivada2023dall}.

\subsection{Task 3: Conceptual Quantitative Reasoning}
In this task we evaluate models on prompts that require a conceptual understanding of objects and their parts and are thus more challenging compared to the previous tasks.
Notions of ``parts'', ``proportion'' and ``fractions'' tend to be concepts that are harder to acquire for both children and animals~\cite{SHIPLEY1990109,woodruff1981primative}.
For example, when a fork is broken into two parts, three- to four-year-old children count each discrete physical object as a separate fork~\cite{SHIPLEY1990109}. 

\noindent\textbf{Prompt types.} Three prompt types are used in this task (the row ``Quantitative'' in Table~\ref{tab:prompt_types}): the \emph{fractional-simple} category includes notions of a one whole, and basic fractions such as \textonehalf\,and \textonequarter\, (\eg, ``A cake cut into quarters.''); the \emph{fractional-complex} prompts include notion of fractions in relation to another attribute such as color or size (\eg, ``one piece is twice the size of the other''). Finally, inspired by the fork experiment with children~\cite{SHIPLEY1990109}, we include \emph{part-whole} prompt type where an object such as a fork, pencil or a plate is split into pieces.

%% file: tables/prompt_examples.tex
\begin{tabular}{c l l}
\toprule
\textbf{Task}  & \textbf{Prompt Type}   & \textbf{Prompt Template / Example Prompts} \\ \midrule
\parbox[t]{2mm}{\multirow{20}{*}{\rotatebox[origin=c]{90}{Exact}} }          & numeric-simple          & \begin{tabular}[c]{@{}l@{}} <num> <noun> \\ \emph{One pistachio.}\\ \emph{7 kangaroos.} \end{tabular} \\ \cmidrule{2-3} 
                                                  & numeric-sentence        & \begin{tabular}[c]{@{}l@{}}
                                                  There <verb> <num> <noun> in this image. \\ \emph{There is 1 fish.}\\ \emph{An image showing 1 fish.}\end{tabular}  \\ \cmidrule{2-3} 
                                                  & 2-additive              & \begin{tabular}[c]{@{}l@{}}
                                                  <num1> <noun1> and <num2> <noun2>.
                                                  \\ \emph{3 durians and three bonsais.}\\
                                                  \emph{Four axolotls and three cookies.}\\\end{tabular} \\ \cmidrule{2-3}
                                                  & 2-additive-color        &  \begin{tabular}[c]{@{}l@{}}
                                                  <num1> <color> <noun1> and <num2> <color> <noun2>.\\
                                                  \emph{Four red cats and four red mushrooms.}\\
                                                  \emph{One black cat and 1 black apple.}\end{tabular} \\ \cmidrule{2-3} 
                                                  & 3-additive              &  \begin{tabular}[c]{@{}l@{}}
                                                  <num1> <noun1>, <num2> <noun2> and <num3>  <noun3>.\\
                                                  \emph{2 parsnips, 3 coconuts and one seahorse.}\\
                                                  \emph{Two manatees, 2 burgers and 1 okra.}\\\end{tabular}     \\ \cmidrule{2-3} 
                                                  & attribute-color         &  \begin{tabular}[c]{@{}l@{}}
                                                  <num> <color> <noun>.\\
                                                  \emph{One black koala.}\\ \emph{4 green bottles.}\end{tabular}       \\ \cmidrule{2-3}
                                                  & attribute-spatial       &  \begin{tabular}[c]{@{}l@{}}
                                                  There <verb> <num1> <noun1> <rel> <num2> <noun2>.\\
                                                  \emph{There are three cats above one manatee.}\\
                                                  \emph{There are three coconuts to the left of 5 burgers.}\end{tabular}       \\ 
                                                        \midrule
\parbox[t]{2mm}{\multirow{5}{*}{\rotatebox[origin=c]{90}{Approx.}}}        & approx-1-entity           &  \begin{tabular}[c]{@{}l@{}}
An image of <noun>. There <verb> <quant> <loc>.\\
\emph{An image of a shelf. There are many books on the shelf.}\\
\emph{An image of a watermelon. There are no seeds in the watermelon.}\end{tabular}       \\ \cmidrule{2-3}
                                                  & approx-2-entity         &  \begin{tabular}[c]{@{}l@{}}
\makecell[l]{An image with some <noun1> and <noun2>. There <verb> <quant> \\ \,<noun1> <quant> <noun2>.}\\
\makecell[l]{\emph{An image with some tables and some pencils. There are fewer tables}\\ \, \emph{than pencils.}}\\ 
\makecell[l]{\emph{An image with some bottles and some apples. There are as many bottles}\\ \, \emph{as apples.}}\end{tabular}      \\ \midrule
\parbox[t]{2mm}{\multirow{9}{*}{\rotatebox[origin=c]{90}{Quantitative}}} & fractional-simple           &  \begin{tabular}[c]{@{}l@{}}
There <verb> <num> <noun> and <frac> of another <noun>.\\
\emph{A pizza cut into 3 slices.}\\ 
\emph{There are three apples and quarter of another apple on the table.}\end{tabular}         \\ \cmidrule{2-3}
                                                  & fractional-complex      &  \begin{tabular}[c]{@{}l@{}}
A <noun> is cut into <num> pieces. One piece is twice the size of the other.\\
\emph{An image of a pencil where one half of it is red and the other half is blue.}\\
\emph{A banana is cut into two pieces. One piece is twice the size of the other.}
 \end{tabular}   \\\cmidrule{2-3}
                                                  & part-whole              &    \begin{tabular}[c]{@{}l@{}}
There <verb> <num> <noun>, but one <noun> is broken into two pieces.\\
\emph{There are 2 forks on the table, but one fork is broken into two pieces.}\\
\emph{There are 5 pencils on the table, but one pencil is broken into two pieces.}
                                                  \end{tabular}      
                                                        \\\bottomrule
\end{tabular}%

%% file: 05-human-annotation.tex
\section{Human Annotations of Images}
\begin{figure}[t]
    \centering
    \includegraphics[width=\linewidth]{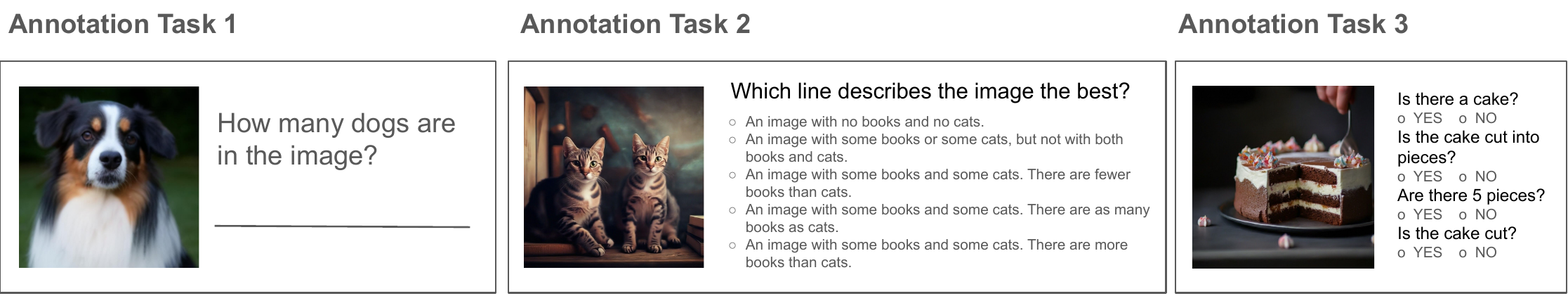}
    \caption{Three types of annotation templates used to collect data for the evaluation of text-to-image models on three numerical reasoning tasks.}%
    \label{fig:annotation_tasks}  
\end{figure}

We evaluate text-to-image models based on annotations collected from study participants who completed three different tasks, where each task corresponds to one of the three numerical reasoning tasks discussed in Section~\ref{sec:curating_prompts}. An example template used for each annotation task is shown in Figure~\ref{fig:annotation_tasks}.

To evaluate models on \textbf{Task 1}, we ask participants to count the number of objects in a generated image. Specifically, we pair each generated image with an automatically generated question \texttt{How many <obj> are in the image?}, where \texttt{<obj>} refers to the noun in the source prompt used to generate the image. We ask one question for each noun in the prompt.
Participants responded with a number representing the count of objects entered in free-form text format; where relevant, they were instructed to provide number ranges (\eg 1--2, 2.5) or ``10+'' labels (see Appendix~\ref{app:annotator_instructions} for detailed annotation instructions).

For \textbf{Task 2}, an image is paired with 3--5 lines of text describing the image. The number of lines depends on the original prompt, for \emph{approx-1-entity} there were three lines, and for \emph{approx-2-entity} there were 5 lines.
Participants were asked to select the line that describes the image the best. The suggested lines were derived from the text prompt used to generate the image (see Figure~\ref{fig:annotation_tasks} for an example).

Finally, to evaluate models on \textbf{Task 3}, we generate a series of short questions covering the words in the prompt words using automatic question generation approach based on Davidsonian Scene Graph (DSG)~\cite{cho2023davidsonian}. The questions appeared next to the image.
DSG generates prompt-specific questions so that the answer to all of them should be ``yes'' if the model accurately depicted contents of the prompt. In analyses, we excluded all numerically irrelevant questions (\ie \texttt{Is this an image?}, \texttt{Is there a table?}, \texttt{Are/is <obj> on the table?}). There are between two and six generated questions per each prompt.

We do not show the original prompt (used to generate a given image) to participants in any of the tasks, as we want to collect unbiased estimates of generated quantities.
Example screenshots showing the web interface participants were seeing for each annotation task are available in Figures~\ref{fig:screenshot1}-\ref{fig:screenshot3}.
Twenty-five participants have been recruited through a crowd-sourcing platform and provided informed consent to participate in the study. Our institution's independent ethical review committee reviewed and approved details of the study design, including working conditions and minimum hourly compensation of £15. 
Each image was annotated by five participants, and we observe a high level of inter-annotator agreement as in over 96\% of cases at least 3/5 annotators provided the same rating (see Appendix~\ref{app:data_analysis} for details).

\noindent\textbf{Processing human annotations.}
Number annotations for object counts in Task 1 are first preprocessed by removing typos and standardizing the format (see Appendix~\ref{app:data_analysis} for details on text preprocessing). 
The mode of the five numbers (\ie the most frequent number) for each image--question pair is considered as the numerical ``label'' (\ie model response) used to calculate model accuracy. 
When calculating accuracy, the ground-truth number is the original number in the text prompt used to generate the image. If the two numbers match, the accuracy is ``1'', otherwise it is ``0''.

In Task 2 each original prompt template is encoded as a number (\eg, ``\texttt{no <X>}'' would be encoded as ``0'', ``\texttt{many <X>}'' as ``4'' etc., see Appendix~\ref{app:encoding_scheme_t2}) taken to be the ground-truth number. 
Each participant response, a radio-button selection, is also encoded as a number on the same scale.
If the two numbers match, the accuracy is ``1'', otherwise it is ``0''.

In Task 3 the ``yes''/``no'' responses given by participants are coded as binary values 1/0 representing accuracy on the question.
We aggregate all binary responses and to obtain the average accuracy for a given prompt--image pair.

%% file: 06-results.tex
\section{Evaluating Text-to-Image Models}
\label{sec:results}

\noindent\textbf{Text-to-image models.} We study 12 different text-to-image models across five  different model families: \dalle~\cite{betker2023improving}, Midjourney, Imagen-based models~\cite{saharia2022photorealistic,vasconcelos2024greedy}, Muse~\cite{pmlr-v202-chang23b} and Stable Diffusion (SD) models~\cite{podell2023sdxl,Rombach2022CVPR,rombach2022high}. 
The models we evaluate cover a wide span of model architectures, including pixel-based (e.g. Imagen) and latent-based (\eg Muse, SD) models. Specifically, the Muse family uses a different generative approach based on predicting masked image patches.

Models within each family may differ in their size, architecture and training data. For models that have been trained on internal data sources we use letters of the alphabet to refer to earlier versions of models (\ie \imagenA is an earlier version of \imagenB, and \imagenD is a recent version referred to as Vermeer~\cite{vasconcelos2024greedy}). 
Generally, the core components of earlier Imagen models have fewer parameters compared to later models (\eg 600M for \imagenA, 2B for \imagenB and \imagenC, and 8B for \imagenD). Imagen A, B and C models have been trained on internally curated versions of WebLI dataset~\cite{chen2022pali}, while \imagenD has been trained on CC12M~\cite{changpinyo2021conceptual}. In all our experiments, for each model we generate five images using five different seeds.

\clearpage
\begin{table}[t]
    \centering
    \small
    \caption{Per task accuracy and the standard error of the mean, with percentage point difference from the baseline in brackets. Accuracy of the best performing model on a task is highlighted in bold, and that of the best model within a family is underlined.}
    \include{./tables/results_all_tasks}
    \label{tab:res1-table}
\end{table}

\begin{figure}[t]
    \centering
    \begin{minipage}{0.45\textwidth}
        \centering
        \vspace{1cm}
        \includegraphics[width=\textwidth]{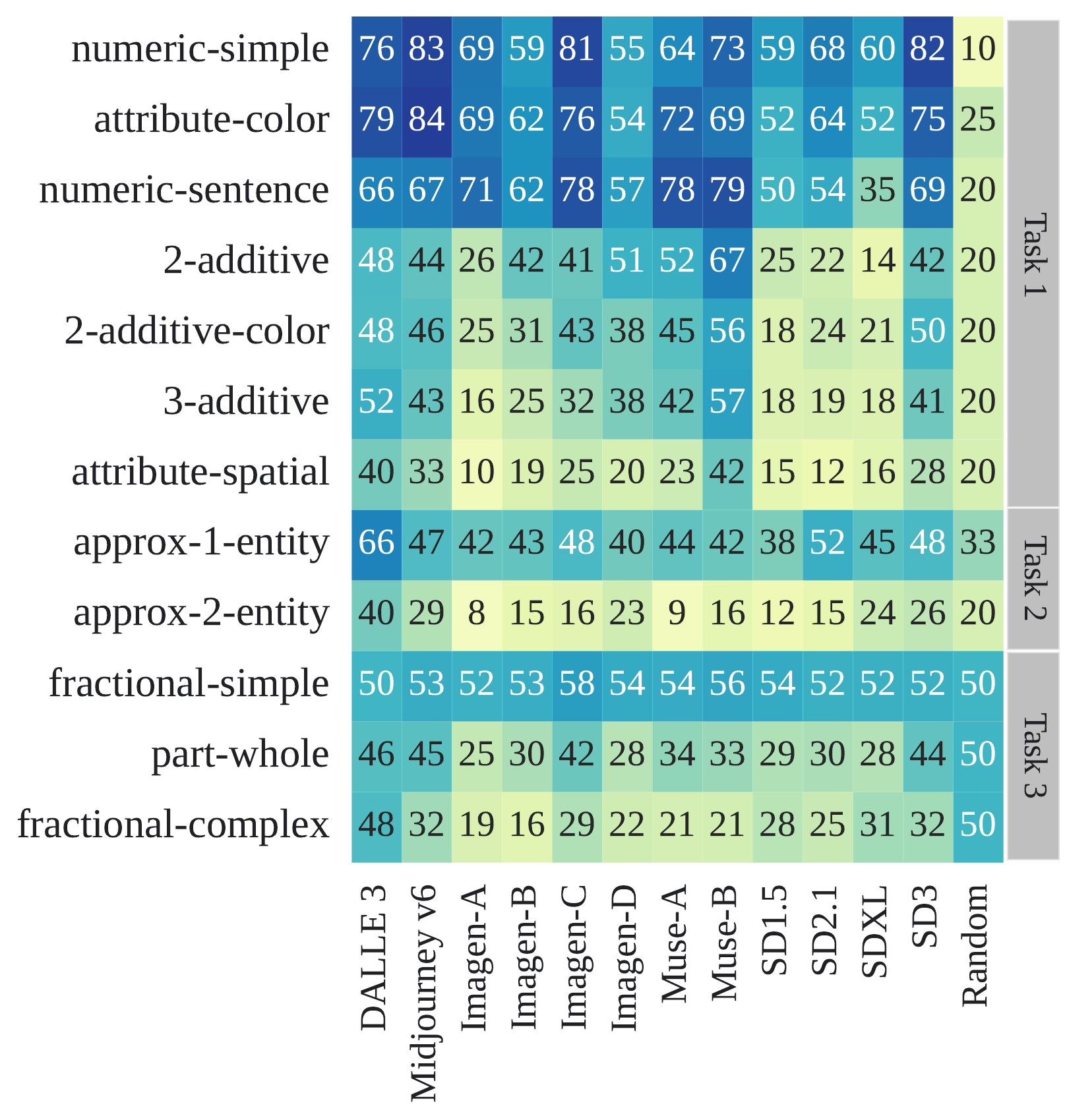}
        \caption{Accuracy of models on each prompt type for a subset of prompts that contain small numbers (\ie 1--4) and a smaller subset of nouns.}
        \label{fig:big_image}
    \end{minipage}\hfill
    \begin{minipage}{0.5\textwidth}
        \centering
        \includegraphics[width=\textwidth]{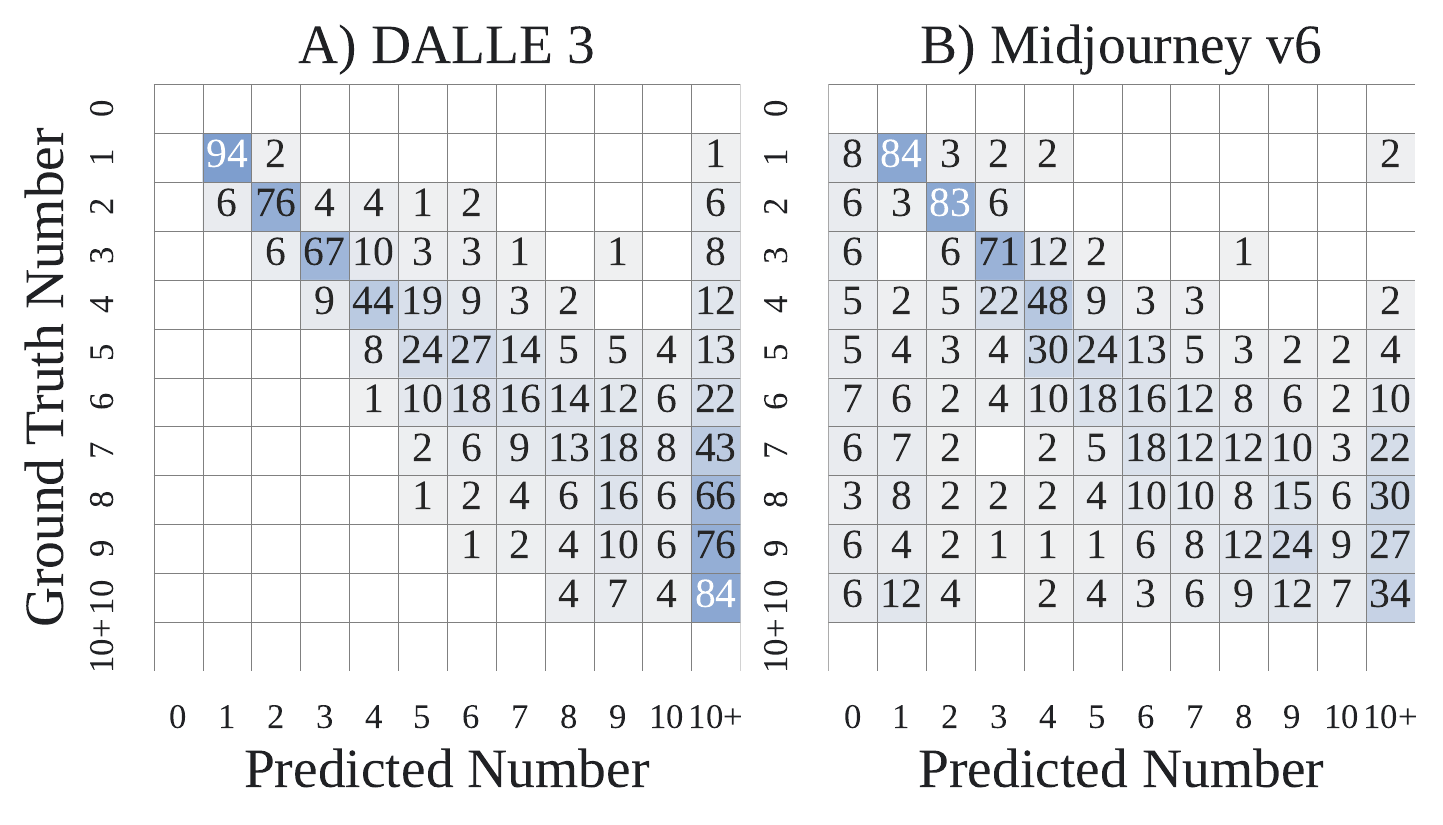}
        \label{fig:small_image1}
        \vspace{-.4cm}
        \includegraphics[width=\textwidth]{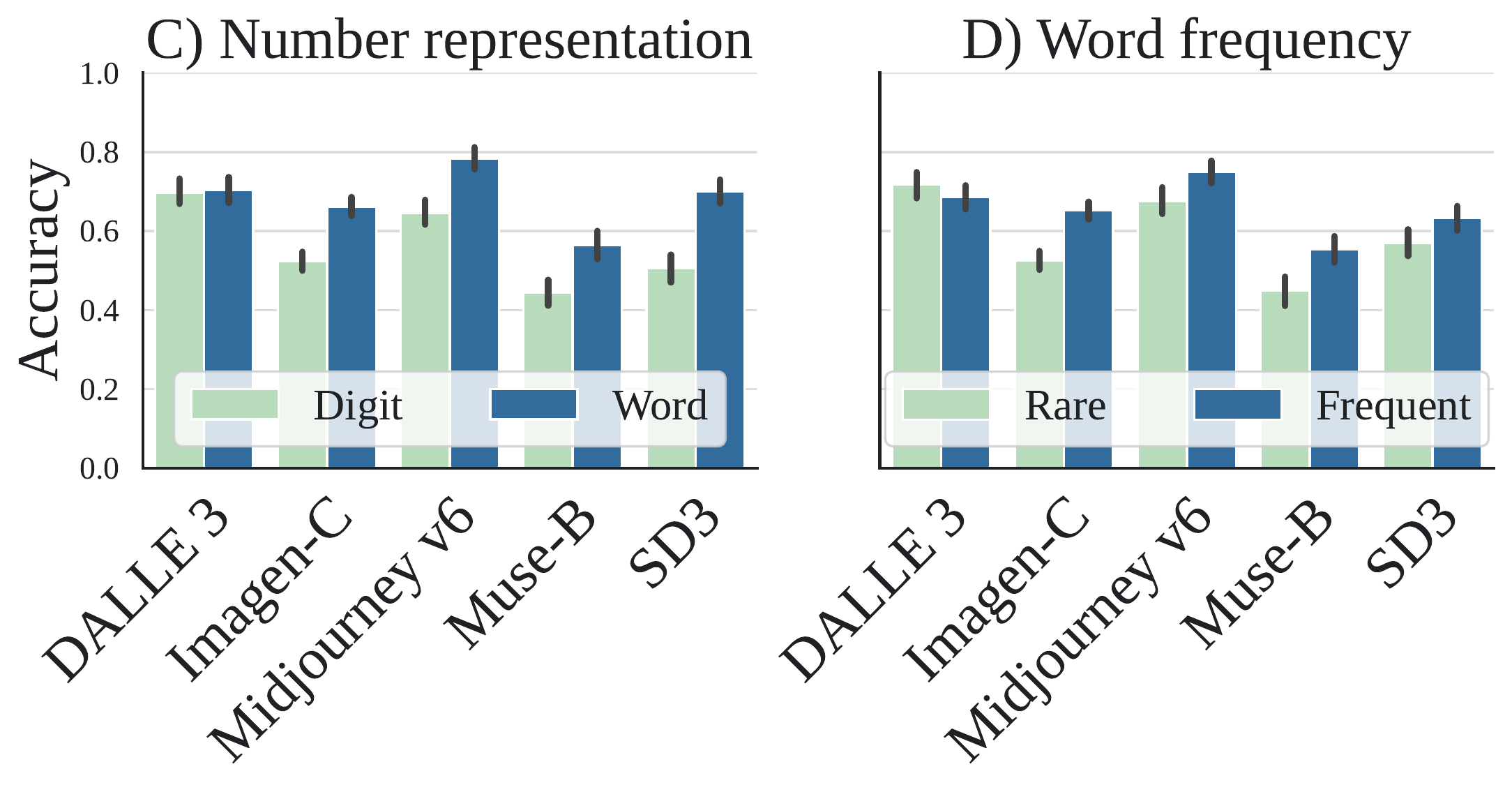}
        \caption{
        Top:  The confusion matrices for A) \dalle and B) \midjourney on \emph{numeric-simple} prompts. Bottom: The effect of C) number representation and D) word frequencies in Task 1. 95\% bootstrap confidence intervals are shown.}
        \label{fig:small_image2}
    \end{minipage}
    \label{fig:arrangement}
\end{figure}
\clearpage

\subsection{Task 1: Exact Number Generation}
\label{sec:task1}

{\bf Number magnitude.} We find that number magnitude strongly affects model performance in number generation, consistent with existing work in the multimodal and generative models~\cite{rane2024can,paiss2023teaching}.
Figure~\ref{fig:small_image2} shows normalized confusion matrices for \dalle and \midjourney, the two strongest models on this task.
Even for \dalle, the accuracy drops substantially with each successive number (18 \pp decrease for 1 $\rightarrow$ 2, 9 \pp for 2 $\rightarrow$ 3, and 23 \pp for 3 $\rightarrow$ 4). All models have a tendency to overestimate numbers (\ie they tend to depict a higher number of entities than what is specified in the prompt), as shown by the non-zero entries over the ``10+'' label on the x-axis in the figure. With the exception of \dalle, we also see underestimation, including instances where models fail entirely to generate the entity in the prompt (non-zero entries over the ``0'' label).

{\bf Number representation (digits vs. words).}
Based on \emph{numeric-simple} prompts and the small number range where models have higher accuracy (1--4), we find that 10 out of 12 models are significantly more accurate when numbers in text prompts are represented with words as opposed to digits, with the exception of \dalle and \sdone, where there was no significant difference.
Figure~\ref{fig:small_image2}C) shows accuracy for a subset of best performing models in each family for Task 1.

{\bf Noun frequency (frequent vs. rare words).}
Nine out of 12 models were more accurate when the nouns associated with numbers in text prompts were frequent words, as opposed to rare words, for \emph{numeric-simple} prompts.
The only exceptions were \dalle, \imagenD and \sdxl, where there was no significant difference in accuracy.
While this finding may indicate that these three models are better at abstracting the notion of a number as opposed to memorizing frequently occurring number-noun combinations, we note that this finding is based on a small set of words $(N=40)$ that may or may not generalize for a greater sample size. The comparison of accuracies for a subset of models is shown in Figure~\ref{fig:small_image2}\.D). 

{\bf Prompt structure.}  Compositional prompts types with more than one number-noun combination, such as \emph{2-additive} and \emph{3-additive}, were on average more difficult compared to prompts with a single noun and a number (\ie \emph{numeric-simple}) (\cf Figure~\ref{fig:big_image}).
With the exception of \dalle, performance of all other eleven models was also significantly lower when numbers appeared in the same context with \textbf{spatial relationships} as in \emph{attribute-spatial} prompts, as those prompts were significantly harder when compared to \emph{2-additive} prompts that did not include spatial relationships. Adding \textbf{color} terms to numbers significantly reduced accuracy in \imagenC, \sdone and \sdthree models.
These results indicate that prompt complexity, such as including additional prepositions or numbers in the text, can dramatically impair the ability of models to correctly generate images containing even a small numbers of entities.
In Section~\ref{sec:qual_failures} in the Appendix we discuss different types of model failures in Task 1, and in Section~\ref{app:task1_qual} we discuss the occasional cases of disagreements in human annotations with several qualitative examples.

\subsection{Task 2: Approximate Number Generation and Zero}
\begin{wrapfigure}{R}{0.4\linewidth}
  \centering
    \vskip-4em
    \includegraphics[width=0.925\linewidth]{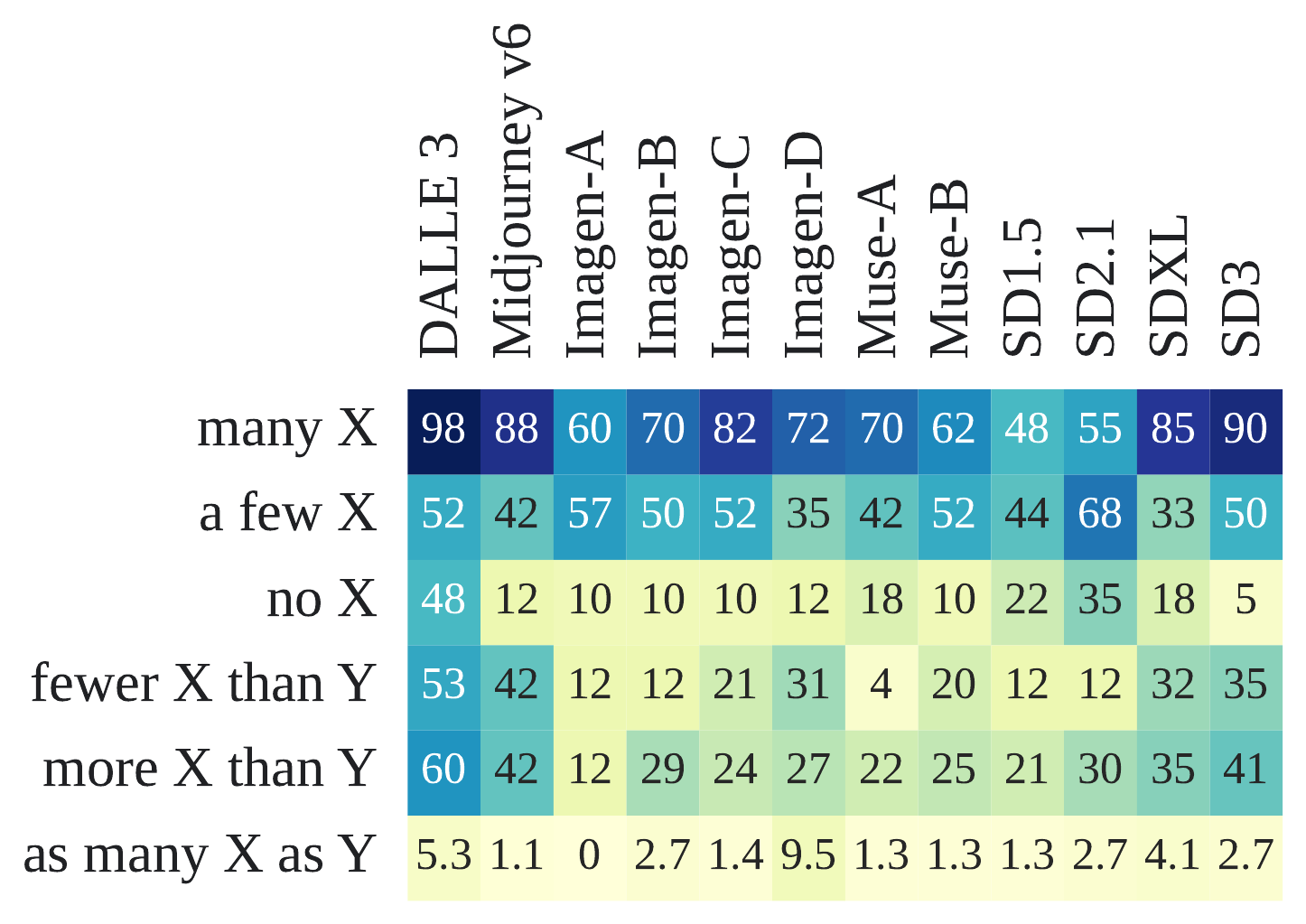}
  \caption{Accuracy for \emph{approx-1-entity} and \emph{approx-2-entity} prompts.}
  \label{fig:res-estimation}
\end{wrapfigure}
When testing for approximate number generation, similarly to Task 1, we observe that prompts with fewer entities (\ie \emph{approx-1-entity})  are on average easier than prompts with two entities (\ie \emph{approx-2-entity}), as seen in Figure~\ref{fig:big_image}.
Figure~\ref{fig:res-estimation} shows a further breakdown of accuracy per specific prompt template. For \emph{approx-1-entity} (first 3 rows containing only ``X''), Eleven out of 12 models were the most accurate in generating ``many'' objects, and all models were least accurate in generating images with zero objects (\ie ``There is/are no X'').
For \emph{approx-2-entity} (last 3 rows, with both ``X'' and ``Y''), 8 out of 12 models have highest accuracy when generating images from prompts of the ``more X than Y''-type and lowest accuracy when generating ``as many X as Y''. Our results highlight that understanding of linguistic quantifiers, even as simple as the word ``no'', can still be challenging for generative models. Section~\ref{app:task2_qual} in the Appendix shows qualitative examples including images and annotations in Task 2.

\subsection{Task 3: Conceptual Quantitative Reasoning}
\label{sec:results_task3}

Task 3 is the hardest, as most models perform close to or below the random chance baseline (\cf~Table~\ref{tab:res1-table}).
 All twelve models had the highest accuracy on \emph{fractional-simple} prompts, followed by \emph{part-whole} and \emph{fractional-complex} for 10 out of 12 models, consistent with intuitions regarding the prompt difficulty. 
The last column in Figure~\ref{fig:results-task1-examples} shows images for a prompt in the \emph{fractional-simple} category, where many models failed to correctly depict ``one apple and half of another apple''. Further examples, including images and annotations, are shown in Table~\ref{fig:app_exampes_task3}.

However, evaluating models on Task 3 prompts was more challenging due to the question-based method used---for some prompts, the DSG method generates only two relevant questions out of which one asks about the presence of an object in the image  (\eg \emph{Are there apples?}) and thus answering ``yes'' yields the baseline of 50\%, without necessarily capturing the nuance between whole apples, sliced apples or thirds/quarters of an apple. Further methodological challenges related to evaluation of models on this task are discussed in Appendix~\ref{app:model_eval}.

%% file: tables/results_all_tasks.tex
\begin{tabular}{llll}
\toprule
 & \parbox{3cm}{\centering \textbf{Task 1} \\ Exact Number Generation} & \parbox{3cm}{\centering \textbf{Task 2} \\ Approximate Number Generation and Zero} & \parbox{3cm}{\centering \textbf{Task 3} \\ Conceptual Quantitative Reasoning} \\
\midrule
\dalle & $\bm{45.2}\pm0.5\,(\textcolor{teal}{+35.2\%})$ & $\bm{48.7}\pm2.7\,(\textcolor{teal}{+24.1\%})$ & $48.8\pm1.1\,(\textcolor{orange}{-1.2\%})$ \\
\midrule
Midjourney v6 & $42.5\pm0.5\,(\textcolor{teal}{+32.5\%})$ & $35.0\pm2.3\,(\textcolor{teal}{+10.4\%})$ & $48.9\pm1.1\,(\textcolor{orange}{-1.1\%})$ \\
\midrule
Imagen-A & $26.3\pm0.4\,(\textcolor{teal}{+16.3\%})$ & $20.0\pm2.2\,(\textcolor{orange}{-4.6\%})$ & $41.1\pm1.3\,(\textcolor{orange}{-8.9\%})$ \\
Imagen-B & $27.0\pm0.4\,(\textcolor{teal}{+17.0\%})$ & $24.6\pm2.3\,(\textcolor{teal}{+0.0\%})$ & $42.9\pm1.4\,(\textcolor{orange}{-7.1\%})$ \\
Imagen-C & $\underline{34.9}\pm0.4\,(\textcolor{teal}{+24.9\%})$ & $27.0\pm2.4\,(\textcolor{teal}{+2.4\%})$ & $\bm{\underline{50.6}}\pm1.2\,(\textcolor{teal}{+0.6\%})$ \\
Imagen-D & $28.5\pm0.4\,(\textcolor{teal}{+18.5\%})$ & $\underline{28.7}\pm2.4\,(\textcolor{teal}{+4.0\%})$ & $43.8\pm1.3\,(\textcolor{orange}{-6.2\%})$ \\
\midrule
Muse-A & $34.8\pm0.4\,(\textcolor{teal}{+24.8\%})$ & $21.0\pm2.2\,(\textcolor{orange}{-3.6\%})$ & $45.1\pm1.2\,(\textcolor{orange}{-4.9\%})$ \\
Muse-B & $\underline{39.8}\pm0.5\,(\textcolor{teal}{+29.8\%})$ & $\underline{24.6}\pm2.3\,(\textcolor{teal}{+0.0\%})$ & $\underline{46.2}\pm1.2\,(\textcolor{orange}{-3.8\%})$ \\
\midrule
SD1.5 & $20.3\pm0.4\,(\textcolor{teal}{+10.3\%})$ & $20.6\pm2.2\,(\textcolor{orange}{-4.0\%})$ & $44.6\pm1.2\,(\textcolor{orange}{-5.4\%})$ \\
SD2.1 & $25.8\pm0.4\,(\textcolor{teal}{+15.8\%})$ & $27.9\pm2.4\,(\textcolor{teal}{+3.3\%})$ & $43.5\pm1.1\,(\textcolor{orange}{-6.5\%})$ \\
SDXL & $22.8\pm0.4\,(\textcolor{teal}{+12.8\%})$ & $31.2\pm2.5\,(\textcolor{teal}{+6.6\%})$ & $43.8\pm1.1\,(\textcolor{orange}{-6.2\%})$ \\
SD3 & $\underline{40.0}\pm0.5\,(\textcolor{teal}{+30.0\%})$ & $\underline{33.9}\pm2.6\,(\textcolor{teal}{+9.3\%})$ & $\underline{47.8}\pm1.0\,(\textcolor{orange}{-2.2\%})$ \\
\midrule
Random Chance & $10.0$ & $24.6$ & $50.0$ \\
\bottomrule
\end{tabular}

%% file: 07-beyond-benchmark.tex
\section{Measuring What Counts: Challenges in Evaluation of Numerical Reasoning}
\label{sec:challenges}

\begin{wraptable}{r}{0.45\linewidth}
\vspace{-12pt}
\small
\caption{
The ability of auto-metrics to discriminate between correct and incorrect image generations. ($\checkmark$ indicates $p<0.05$ using Mann-Whitney U test)}
\resizebox{\linewidth}{!}{
\begin{tabular}{l|ccccc}
\toprule
          & CLIP       & TIFA    & Gecko      & DSG         & VNLI \\ \midrule
\dalle    & =           & =         & \checkmark & \checkmark   & \checkmark \\
\midjourney  & =   & \checkmark  & \checkmark & \checkmark   & \checkmark \\
\imagenC    & =  & \checkmark  & \checkmark & \checkmark   & \checkmark \\
\museB   & \checkmark & \checkmark & \checkmark & \checkmark & \checkmark \\
\sdthree   & = & \checkmark &  \checkmark  & \checkmark & \checkmark \\
 \bottomrule
\end{tabular}
}
\label{tab:autoeval}
\end{wraptable}

We explore the utility of \geckonum for two related areas of research, developing automatic evaluation metrics and evaluating vision--language models (VLMs) with respect to numerical reasoning

\noindent\textbf{Evaluating auto-eval metrics.}
Developing metrics that can reliably replace human evaluation is an active area of research. In particular, a recent line of work proposes \emph{auto-eval metrics} for measuring text-to-image alignment by using pretrained language and/or VLMs ~\cite{hessel2021clipscore,hu2023tifa,wiles2024revisiting,cho2023davidsonian,yarom2024you}.
Here, we evaluate such auto-eval metrics on their ability to capture exact number generation by examining how well they distinguish between correctly and incorrectly generated images for \emph{numeric-simple}  prompts containing small numbers (\ie 1--4) in Task 1.

We consider five different auto-eval metrics: CLIPscore~\cite{hessel2021clipscore}, a metric based on computing similarity between the text and an image, question-answering (QA) based metrics such as TIFA~\cite{hu2023tifa}, Gecko~\cite{wiles2024revisiting}, DSG~\cite{cho2023davidsonian}, as well as VNLI~\cite{yarom2024you}, a metric fine-tuned to predict alignment between images and text. 
We divide images generated by \dalle, \midjourney, \imagenC, \museB, and \sdthree into two groups each based on whether the image was generated correctly or not. Then, we statistically test whether the distributions of scores in the two groups is identical.
If there is a statistically significant difference between the two distributions, and if the correct generations have a higher score, we mark that case with a $\checkmark$ in Table~\ref{tab:autoeval}.
Gecko, DSG and VNLI are able to reliably distinguish between correct and incorrect image generations for all models.
Upon manual inspection of generated questions for each QA method, we observed that Gecko and DSG generate similar questions, while TIFA occasionally generates questions about concepts that do not exist in the prompt.
Following the approach in~\cite{wiles2024revisiting}, we use a Wilcoxon signed-rank test to test whether there is a significant difference in average scores between pairs of models for the same metric. Only VNLI and Gecko were able to correctly order models in pairwise comparisons, in 8/10 and 7/10 cases, respectively.
We also note that this analysis is based on simple prompts which trigger simple questions and thus most of the heavy lifting in performance of an auto-metric is delegated to the underlying VQA model.

\paragraph{Evaluating counting in vision-language models (VLMs).} 
While counting is known to be challenging for VLMs, only a few datasets and benchmarks exist to train and evaluate VLMs on counting~\cite{acharya2019tallyqa,trott2018howmanyqa}.
We explore if \geckonum, which VLMs have never been explicitly trained on, can be used as an evaluation task by curating a VQA benchmark. We evaluate the accuracy of \paligemma~\cite{paligemma}, a state-of-the-art open-source VLM, on counting. Technical details about the setup and more detailed results of experiments in this section are available in Appendix~\ref{app:vqa}.

Comparing the base \paligemma model to a checkpoint that has been fine-tuned for counting on TallyQA (train)~\cite{acharya2019tallyqa}, we find that the fine-tuned model performs better ($73.3{\scriptstyle\pm0.4}\%$) on \geckonum when compared to the base model ($68.4{\scriptstyle\pm0.4}\%$).  
We observe that the base model already performs well on small numbers (up to $4$), but fine-tuning on another counting dataset (\ie, TallyQA) improves accuracy on higher counts (${}\geq 5$), a pattern we also observe with \geckonum.
We also briefly investigate the utility of \geckonum as synthetic training data: we fine-tune base \paligemma on a mixture of TallyQA (train) and images from all \imagen models, and evaluate it on TallyQA (test) and \museB images (see Appendix~\ref{app:vqa} for setup details). We find that including \imagen data does not significantly change performance on TallyQA (test) but vastly improves performance on \museB (by more than $20$ \pp in some cases). This is still true when we test on a set of held-out \museB classes that we removed from the \imagen fine-tuning data. These preliminary results highlight that: 1) training VLMs with synthetic data can improve results on other benchmarks; 2) there is great need for more public high-quality datasets and benchmarks that evaluate counting and numerical reasoning.
\vspace{-.1cm}

%% file: 08-discussion.tex
\section{Discussion and Conclusion}
\label{sec:discussion}
Most work on evaluation of text-to-image models uses generic prompts---manually written or harvested form the Web---as a starting point to generate images. 
For example, recent research has focused on designing comprehensive prompt sets that measure text-to-image alignment~\cite{wiles2024revisiting,hu2023tifa,cho2023davidsonian,huang2024t2i,tuo2023anytext,gokhale2022benchmarking}.

In this work, we propose \geckonum, a benchmark that specifically focuses on measuring numerical reasoning capacity through three tasks: (i) exact and (ii) approximate number generation, and (iii) conceptual reasoning about quantities. We define various prompt types within these tasks to better control for the effect of factors such as the context of the number in the sentence and the sentence structure. 
We use \geckonum to evaluate twelve different text-to-image models from five different model families (\dalle, \midjourney, Imagen, Muse and Stable Diffusion) by collecting human annotations for the generated images.
We find that \dalle has the highest overall accuracy on exact and approximate number generation, while being the least impacted by some prompt manipulations we investigate (\eg numbers represented as words vs. digits, frequent vs rare words); however, its performance on these tasks is still close to or under 50\%. %
Depiction of approximate quantities and the concept of zero was also a weak point in all models. For example, even \dalle would consistently fail to correctly generate ``a watermelon with no seeds'' or ``a cake with no candles''.
We also find that the task requiring reasoning about parts and fractions was challenging for all models as their performance was close to the baseline. 

We used \geckonum to show that only some auto-eval metrics can reliably differentiate between correct and incorrect images on simple numerical prompts---among metrics we tested, our preliminary experiments show that VNLI~\cite{yarom2024you} and Gecko~\cite{wiles2024revisiting} are the only two metrics capable of such differentiation, including the ability to rank models in pairwise model comparisons.
We also demonstrated that our benchmark could be used to study and potentially improve the performance of pretrained vision--language models on counting.

Our approach relies on human annotators to provide counts of objects in an image, as is often the gold standard in evaluation of text-to-image models.
However, such annotation process is laborious, costly and does not scale. 
We expect that the rapid improvement of pipelines and frameworks for evaluation of text-image alignment will reduce the need for manual annotation~\cite{cho2024visual,yarom2024you,gokhale2022benchmarking}.
In addition, we identified three important numerical capabilities for which we have manually designed human annotation templates, but numerical cognition in humans spans a wide set of gradually developing capabilities that might require additional evaluation templates. 
As discussed in Section~\ref{sec:results_task3}, the design of protocols for evaluation of more complex aspects of numerical reasoning remains an important open challenge with rapid advancements in models.
Overall, our results indicate that the current text-to-image models do not form abstract representations of numbers as their ability to reason about numbers is rudimentary: it is limited to depiction of exact, small quantities in images, and models frequently fail to generate approximate quantities and zero. We highlight the effectiveness of numerical reasoning as an open challenge in evaluation, since even in the large-scale training regime, due to the large combinatorial space of numbers, it is challenging to mitigate models' weaknesses with more training data. Future modeling innovations---and not only better training data---might be needed to improve model performance on numerical reasoning.
Our results show that most models are sensitive to the number representation in a given prompt (words vs digits). This showcases another interesting challenge involved in evaluating for numbers:  they can occur in various formats (\eg, dates, phone numbers, or expressions such as ``4k'') and representations. Also, effective ways of tokenizing numbers, again due to the vast space of possible numbers, is an open problem in language modeling \cite{singh2024tokenization} and an interesting direction for future research.

\paragraph{Acknowledgments.} We are grateful to our colleagues at Google DeepMind for advice and feedback that helped improve this work. 
We thank Cristina Vasconcelos for a detailed review of the manuscript, Jason Baldridge for support and useful feedback, Richard Tucker and Aayush Upadhyay for their help and technical support essential in realization of this work, and to Cyrus Rashtchian and Matko Bo\v{s}njak for valuable feedback.

%% file: 0x-checklist.tex
\section*{Checklist}

\begin{enumerate}

\item For all authors...
\begin{enumerate}
  \item Do the main claims made in the abstract and introduction accurately reflect the paper's contributions and scope?
    \answerYes{}
  \item Did you describe the limitations of your work?
    \answerYes{}
  \item Did you discuss any potential negative societal impacts of your work?
    \answerNA{}
      \item Have you read the ethics review guidelines and ensured that your paper conforms to them?
    \answerYes{}
\end{enumerate}

\item If you are including theoretical results...
\begin{enumerate}
  \item Did you state the full set of assumptions of all theoretical results?
    \answerNA{}
	\item Did you include complete proofs of all theoretical results?
    \answerNA{}
\end{enumerate}

\item If you ran experiments (e.g. for benchmarks)...
\begin{enumerate}
  \item Did you include the code, data, and instructions needed to reproduce the main experimental results (either in the supplemental material or as a URL)?
    \answerYes{}{\url{https://github.com/google-deepmind/geckonum_benchmark_t2i}}
  \item Did you specify all the training details (e.g., data splits, hyperparameters, how they were chosen)?
    \answerYes{Section~\ref{sec:challenges}}
	\item Did you report error bars (e.g., with respect to the random seed after running experiments multiple times)?
    \answerYes{}
	\item Did you include the total amount of compute and the type of resources used (e.g., type of GPUs, internal cluster, or cloud provider)?
    \answerYes{}
\end{enumerate}

\item If you are using existing assets (e.g., code, data, models) or curating/releasing new assets...
\begin{enumerate}
  \item If your work uses existing assets, did you cite the creators?
    \answerYes{}
  \item Did you mention the license of the assets?
    \answerYes{}{}
  \item Did you include any new assets either in the supplemental material or as a URL?
    \answerYes{}
  \item Did you discuss whether and how consent was obtained from people whose data you're using/curating?
    \answerYes{}
  \item Did you discuss whether the data you are using/curating contains personally identifiable information or offensive content?
    \answerYes{}
\end{enumerate}

\item If you used crowdsourcing or conducted research with human subjects...
\begin{enumerate}
  \item Did you include the full text of instructions given to participants and screenshots, if applicable?
    \answerYes{See Appendix~\ref{app:annotator_instructions}}
  \item Did you describe any potential participant risks, with links to Institutional Review Board (IRB) approvals, if applicable?
    \answerYes{}
  \item Did you include the estimated hourly wage paid to participants and the total amount spent on participant compensation?
    \answerYes{}
\end{enumerate}

\end{enumerate}

%% file: app_prompts.tex
\section*{Appendices}
\hypersetup{linkcolor=black} 
\printcontents[appendices]{l}{1}{\setcounter{tocdepth}{3}}
\hypersetup{linkcolor=red} 

\renewcommand{\thefigure}{A\arabic{figure}}
\setcounter{figure}{0}

\renewcommand{\thetable}{A\arabic{table}}
\setcounter{table}{0}

\clearpage
\section{Additional Information on Prompts}
\label{app:prompts} 

\begin{table}[h]
    \centering
    \caption{Nouns used in text prompts in Task 1.} 
    \include{./tables/appendix/vocab}
    \label{tab:app_vocabulary}
\end{table}

\begin{wrapfigure}{r}{0.5\textwidth}
    \centering
    \includegraphics[width=0.8\linewidth]{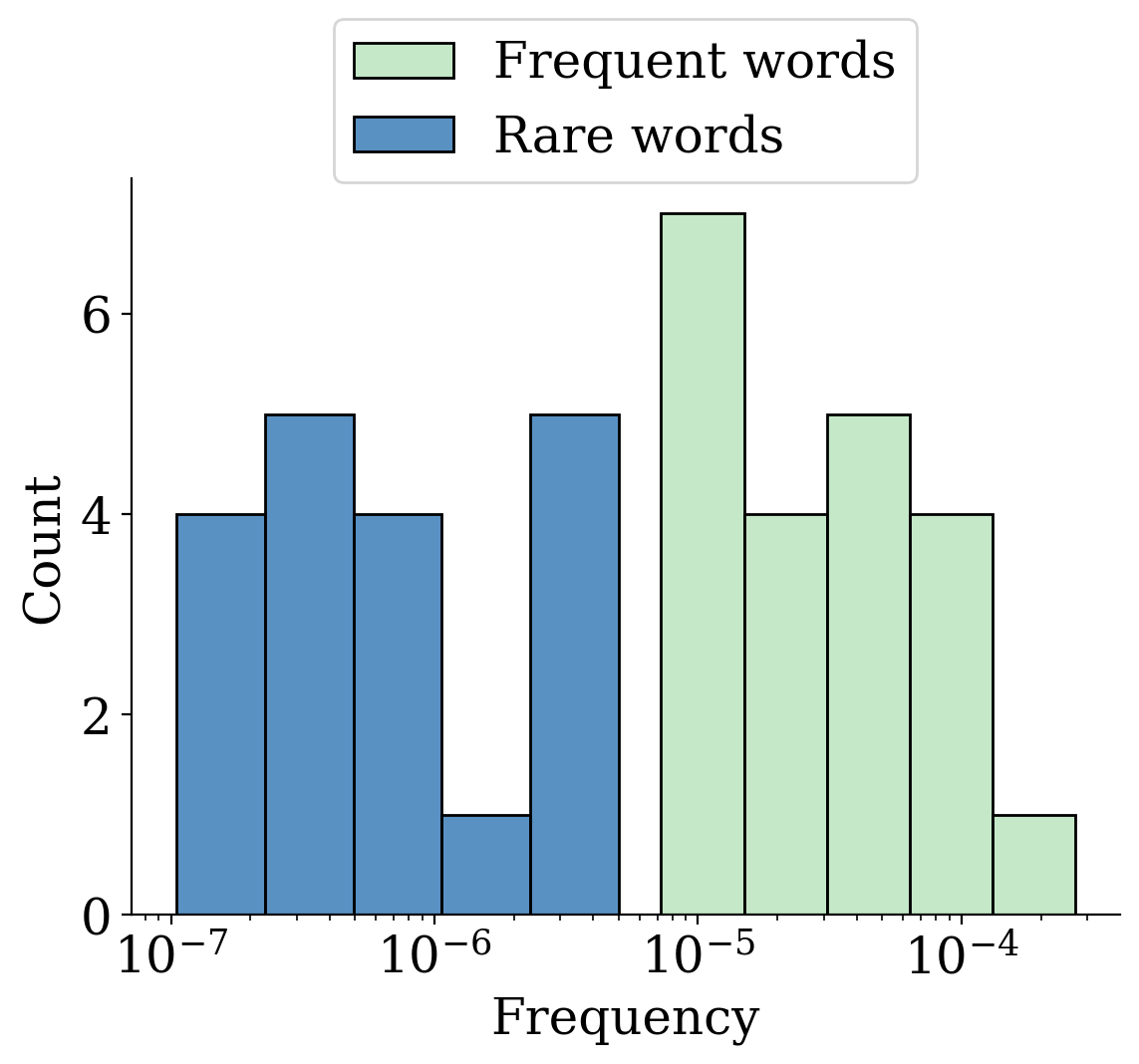}
    \caption{The distribution of frequencies for the nouns used in text prompts in Task 1.}\label{fig:app_frequencies}
\end{wrapfigure}

\subsection{Words and Word Frequencies}
Nouns used in text prompts in Task 1 are selected from the list of 40 words, listed in Table~\ref{tab:app_vocabulary}. The words cover different semantic categories such as everyday objects, food, nature and animals.
Approximately the half of them ($N=21$) are labelled as frequent, and the rest ($N=19$) as non-frequent (\ie rare), as word frequency was a variable we investigated in Section~\ref{sec:task1}.
The distribution of frequencies for those words is shown in Figure~\ref{fig:app_frequencies}, and frequencies are obtained using the Python wordfreq library~\cite{robyn_speer_2022_7199437}  that gathers frequency data based on multiple sources of data such as Wikipedia, OPUS OpenSubtitles 2018~\cite{lison2016opensubtitles2016} and Google Books Ngram, among many others.
For the word `cinnamon stick` we used the word `cinnamon` which was present in the database.
In Tasks 2 and 3 we mostly used the same words as in Task 1, but in Task 3 also added a few additional objects/foods such as ``fork'', ``a loaf of bread'' or ``cake'', where we investigated how well text-to-image models can generate parts of objects or fractions. 

\clearpage
\subsection{The Distribution of Prompt Types in the Benchmark}
\begin{table}[h]
\renewcommand{\arraystretch}{1.3}
    \centering
    \caption{The distribution of prompt types in the benchmark ($N=1386$)}
    \include{./tables/appendix/distr_prompts}
    \label{tab:app_dist_prompts}
\end{table}

\subsection{The Encoding Scheme for Task 2 Answers}
\label{app:encoding_scheme_t2}
\begin{itemize}
    \item 0: An image with no X and no Y. // There are no X.
    \item 1: An image with some X or some Y, but not with both X and Y.
    \item 2: There are fewer X than Y. // There are only a few X.
    \item 3: There are as many X as Y.
    \item 4: There are more X than Y. // There are many X.
\end{itemize}
\clearpage

%% file: tables/appendix/vocab.tex
\begin{tabular}{cc}
\toprule
\textbf{Frequent words} & \textbf{Rare words} \\
\midrule
apple & lychee \\
egg & parsnip \\
burger & samosa \\
cookie & cinnamon stick \\
pizza & axolotl \\
dog & seahorse \\
fish & kangaroo \\
cat & koala \\
ant & manatee \\
fly & mushroom \\
tree & durian \\
leaf & bonsai \\
flower & pistachio \\
coconut & okra \\
olive & crib \\
table & paperclip \\
book & flute \\
bottle & trowel \\
spoon & corkscrew \\
pencil &   \\
chocolate &   \\
\bottomrule
\end{tabular}

%% file: tables/appendix/distr_prompts.tex
\begin{tabular}{llrl}
\toprule
 \textbf{Task} & \textbf{Prompt Type} & \textbf{Number of prompts} & \textbf{Numbers} \\
\midrule
\parbox[t]{2mm}{\multirow{7}{*}{\rotatebox[origin=c]{90}{Task 1}}} 
& numeric-simple & 600 & 1, 2, 3, 4, 5, 6, 7, 8, 9, 10 \\
& attribute-color & 160 & 1, 2, 3, 4 \\
& numeric-sentence & 100 & 1, 2, 3, 4, 5 \\
& 2-additive & 100 & 1, 2, 3, 4, 5 \\
& 2-additive-color & 100 & 1, 2, 3, 4, 5, 6, 7, 8 \\
& 3-additive & 100 & 1, 2, 3, 4, 5 \\
& attribute-spatial & 100 & 1, 2, 3, 4, 5 \\
\midrule
\parbox[t]{2mm}{\multirow{2}{*}{\rotatebox[origin=c]{90}{Task 2}}} 
& approx-1-entity & 24 & no, few, many \\
& approx-2-entity & 45 & fewer, as many as, more \\
\midrule
\parbox[t]{2mm}{\multirow{3}{*}{\rotatebox[origin=c]{90}{Task 3}}} 
& fractional-simple & 36 & 1, 2, 3, 1/2, 1/3, 1/4, 1/5 \\
& part-whole & 15 & 1/2 \\
& fractional-complex & 6 & 1/3 + 2/3, 1/2 \\
\bottomrule
\end{tabular}

%% file: app_results.tex
\section{Additional and Detailed Experimental Results}
\label{app:results}

In this section we show results for all models for experiments in Section~\ref{sec:results}.
In significance tests we use $\alpha=.05$ when rejecting the null hypothesis based on the p-value of a test.
When Chi-squared test is used to compare accuracies  between two groups representing two different conditions (\ie ``digits'' vs ``numerals'', or ``frequent'' vs ``rare'' words), we build a contingency table based on binary accuracy counts for prompt--image pairs in different groups.
In those cases, the null hypothesis is that there is no significant difference in accuracy between the two groups.
We report accuracies as percentages in tables for easier comparison with other values in the manuscript.
As a control condition, \ie to confirm that the variable we investigate indeed explains the differences in results, we also conducted permutation tests where we randomly permuted labels between the two groups under comparison.
Unless indicated differently in the text in those conditions we found that all differences in accuracies were not significant.

\clearpage
\subsection{Task 1: Exact Number Generation}
\subsubsection{Number magnitude}
\begin{figure}[htbp]
    \centering
    \includegraphics[width=.8\linewidth]{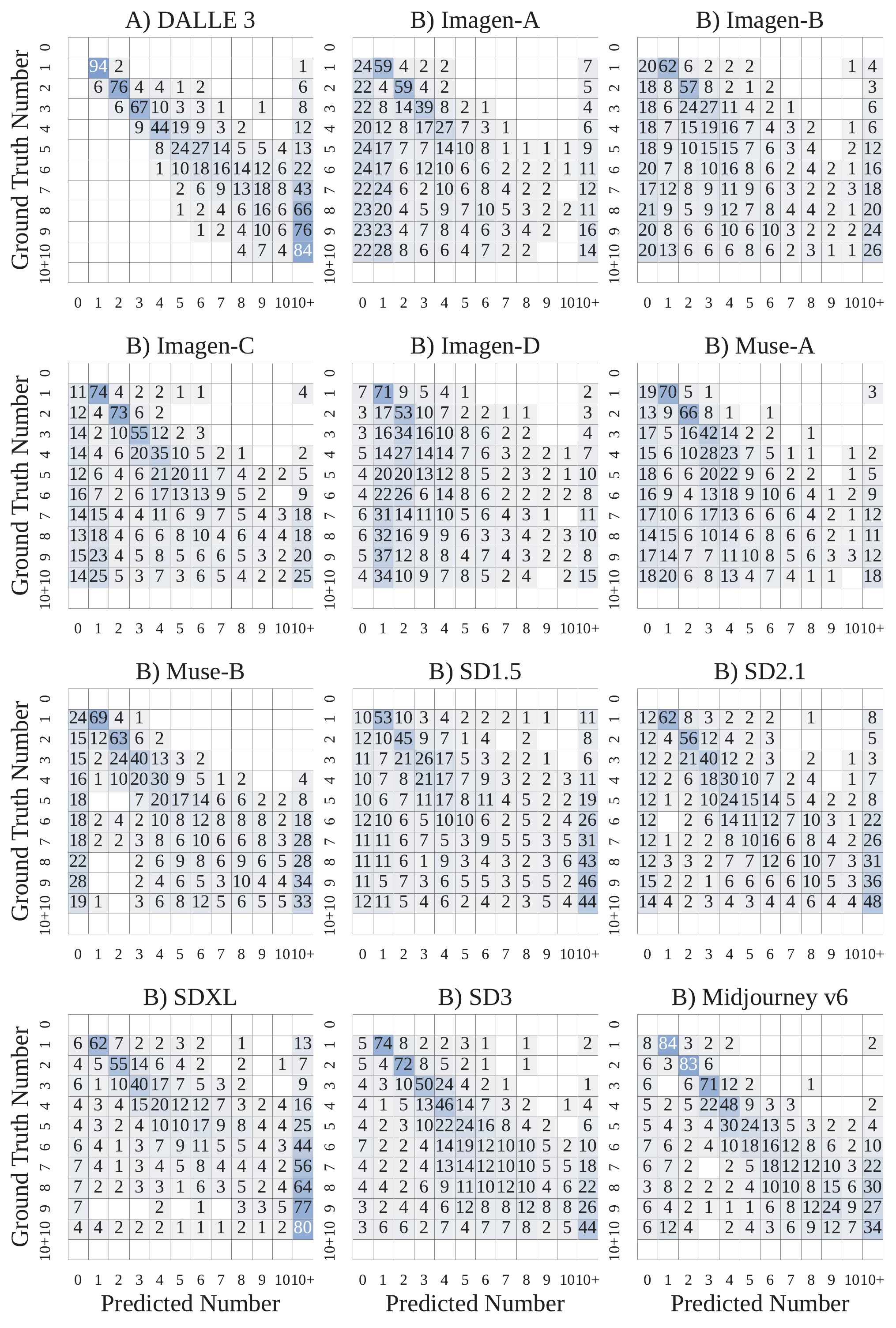}
    \caption{The confusion matrices for all models for \emph{numeric-simple} prompt type. 
    The generated number of objects in an image (as annotated by humans) is on the x-axis, and the ground truth number in the original text prompt is shown on the y-axis. Empty cells denote 0.}
    \label{fig:app_results-magnitude}
\end{figure}

\subsubsection{Number representation and word frequency}
\begin{figure}[H]
    \centering
    \includegraphics[width=\linewidth]{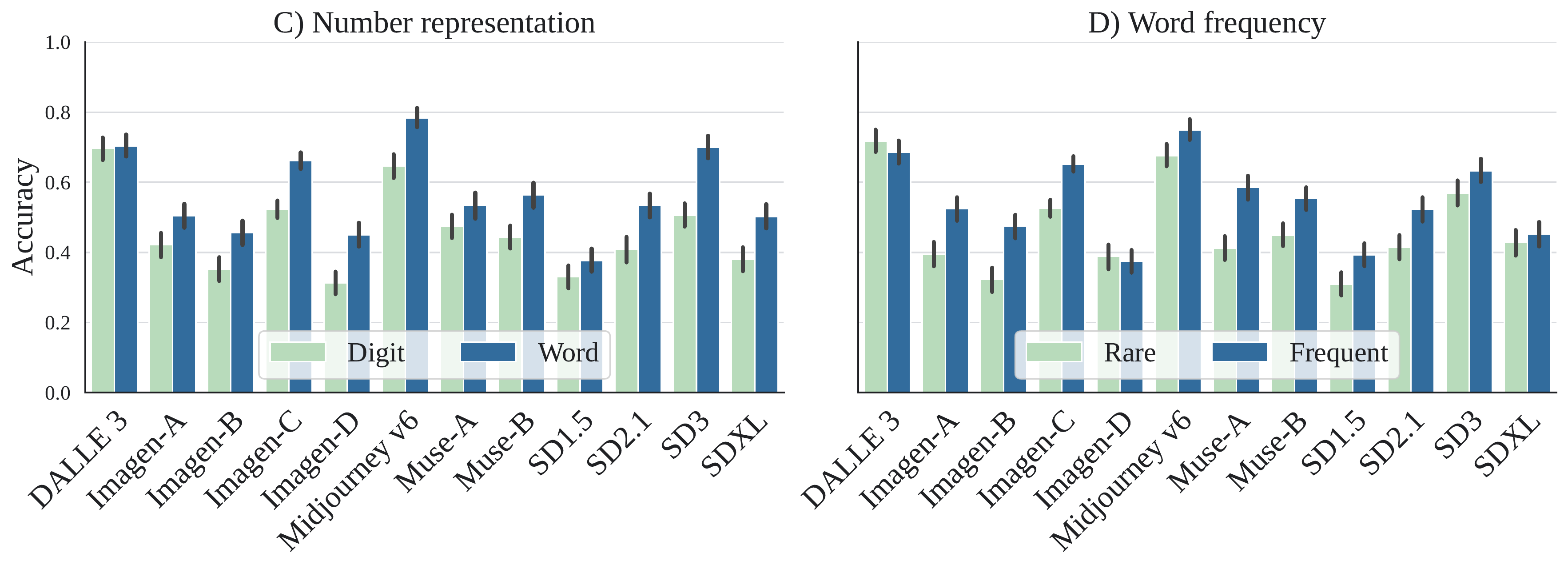}
    \caption{Number representation (A) and word frequency (B) accuracy for \emph{numeric-simple} prompts and all models. Number range: 1--4. 
    Lines on top of bars show 95\% bootstrapped confidence intervals.
    }.
    \label{fig:app_number_rep}
\end{figure}

\begin{table}[h]
\footnotesize
\centering
\caption{Chi-squared test results for comparison of accuracy between prompts with digits ($Acc_D$) and prompts with word numerals ($Acc_W$).
N: Number of samples, $\chi^2$: Chi-squared statistic value, Diff?: whether the difference between two accuracies is significant, $\checkmark$: significant difference, =: insignificant difference.}
\include{./tables/appendix/chi_representation}\label{tab:app_has_numeral}
\quad
\caption{Chi-squared test results for comparison of accuracy between prompts with rare words ($Acc_{R}$) and prompts with frequent words ($Acc_F$).
N: Number of samples, $\chi^2$: Chi-squared statistic value, Diff?: whether the difference between two accuracies is significant, $\checkmark$: significant difference, =: insignificant difference.}
\include{./tables/appendix/chi_freq}\label{tab:app_is_frequent}
\end{table}

\clearpage
\subsubsection{Prompt structure: Additive prompts}
\label{app:additive_results}
\begin{table}[h]
    \centering
    \small
    \caption{Percentage point drop in accuracy for a specific number (1--3) present in different prompt types.}
    \include{./tables/res-additives}
    \label{tab:res-additives}
\end{table}

\begin{figure}[h]
    \centering
    \includegraphics[width=.9\linewidth]{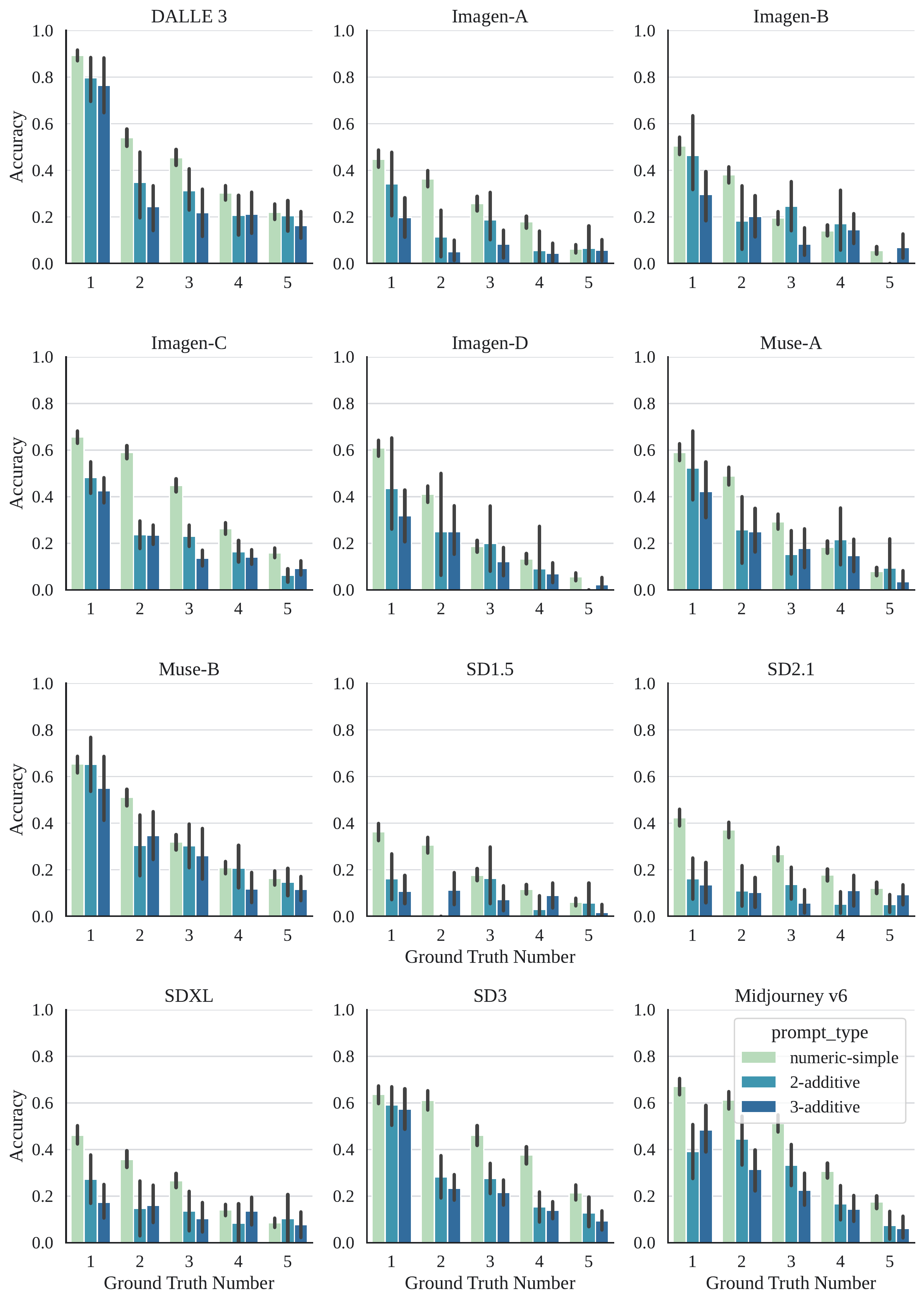}
    \caption{Accuracy per ground truth number in the prompt for different additive prompt types. Lines on top of bars show 95\% bootstrapped confidence intervals.
    }.
    \label{fig:app_additive}
\end{figure}

Figure~\ref{fig:app_additive} shows drops in accuracy when a specific number  occurs in different prompt types.
In addition, Table~\ref{tab:res-additives} shows the exact drops in accuracy for selected models for the numbers: ``1'', ``2'' and ``3''.
While for some models the drop in accuracy for the number ``1'' (\cf the three bars above the x-tick ``1'') is much smaller (\eg \dalle) or non-significant (\museB) compared to accuracy drops for some other models (\eg \imagenA, \imagenC), all models show a substantial and significant drop when ``2'' is present in the prompt (\cf the three bars above the x-tick ``2'').

\clearpage
\subsubsection{Prompt structure: Colors and spatial relationships}
\begin{figure}[h]
    \centering
    \includegraphics[width=.6\linewidth]{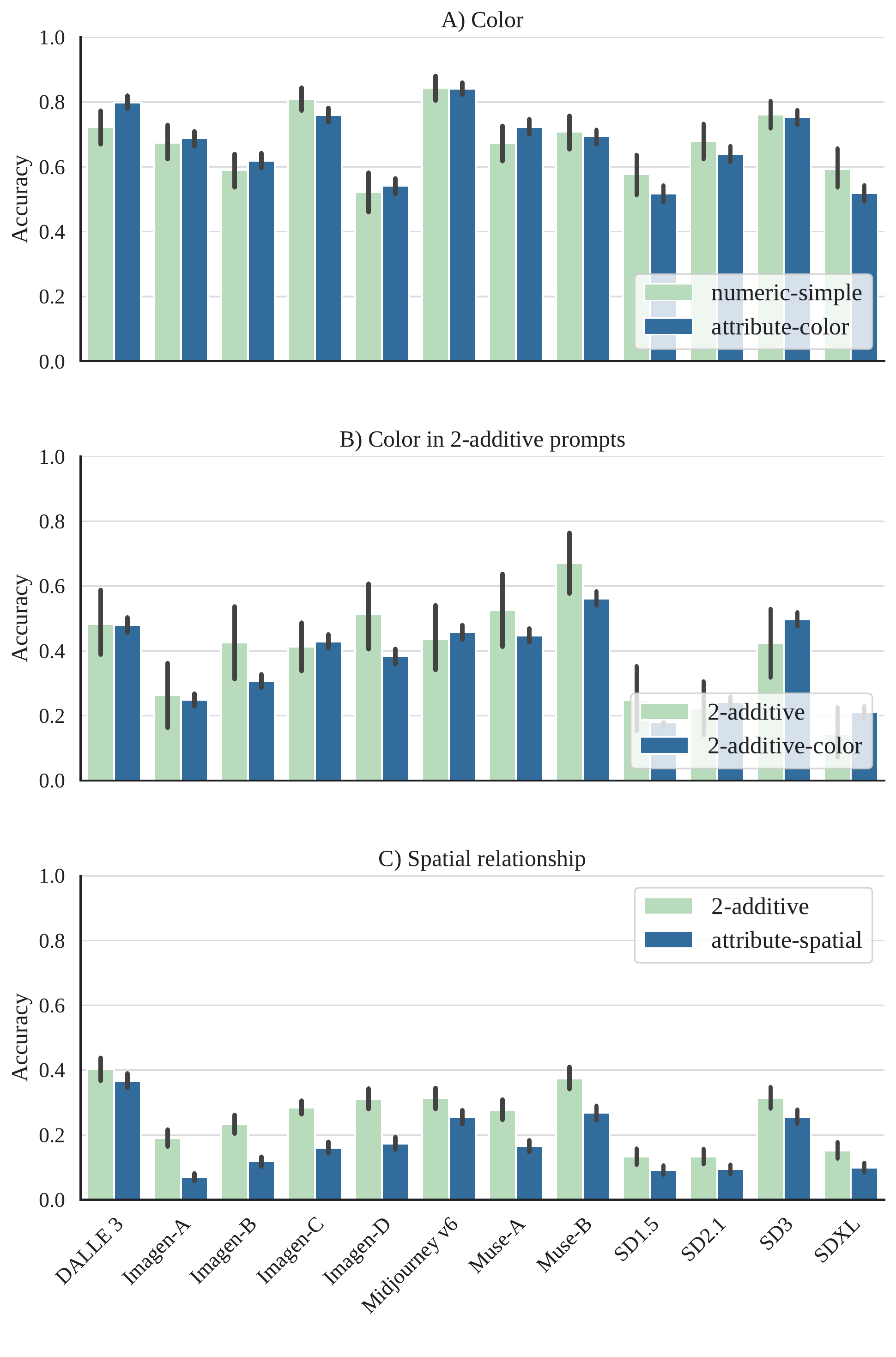}
    \caption{Accuracy in exact number generation when: A) color terms are added to number terms in the prompt, B) color terms are added in 2-additive prompts, and C) spatial relationships are introduced. Lines on top of bars show 95\% bootstrapped confidence intervals.
    }.
    \label{fig:APP_color_spatial}
\end{figure}

To investigate how the addition of color terms and spatial relationships affects accuracy in number generation we ask the following three questions:
\begin{enumerate}
    \item Does adding color adjectives to numbers (such as ``1 red koala'') affect the number generation accuracy?
    \item Does adding color adjectives to numbers in 2-additive prompts (such as ``1 red koala and two green cats'') affect the number generation accuracy?
    \item Does the introduction of spatial relationships between nouns associated with numbers affect the number generation accuracy?
\end{enumerate}

To answer the first question, we split the data into the two groups: the first group contains the subset of data for \emph{numeric-simple} prompts, and the second group the subset of data for \emph{attribute-color} prompts.
We only consider prompts in both groups that contain the same numbers (1--4) and the same words (``cat'', ``apple'', ``koala'', ``bottle'', ``mushroom''), to isolate the effect of adding the color term as opposed to potential confounding factors. 
For example a confounding factor might be the word identity, as a model might be more accurate in generating correct images when the prompt contains the word ``dog'', and if this word exists only in the first prompt type and not in  the second then responses in the first prompt type will on average have higher accuracy that may or may not depend on color terms.
We split the data in a similar way for the other two questions. 
To answer the second question we split the data based on \emph{2-additive} and \emph{2-additive-color} prompts, and to answer the third question we split the data into \emph{2-additive} and \emph{attribute-spatial} prompts. 

We show the average accuracy between groups for each of these conditions in Figure~\ref{fig:APP_color_spatial}, and significance test results based on Chi-squared test in Tables~\ref{tab:app_color}-\ref{tab:app_spatial}.
We see that most differences were not significant when the color term was added, with the exception of \museA in Figure~\ref{fig:APP_color_spatial}A) where the accuracy was actually higher when the color term was added,
and \imagenB and \imagenD in Figure~\ref{fig:APP_color_spatial}B).
In both Figures~\ref{fig:APP_color_spatial}A) and B) we highlight relatively large confidence intervals for \emph{numeric-simple} and \emph{2-additive} groups.
Those two types of prompts were associated with fewer datapoints, compared to the groups they were compared to (\ie \emph{attribute-color} and \emph{2-additive-color}, respectively).
While the average accuracies are much closer in Figure~\ref{fig:APP_color_spatial}A), we speculate that with more samples in Figure~\ref{fig:APP_color_spatial}B) we may see more significant differences for some other models. In contrast, when the prompt contained a spatial relationship, we see significant differences for all models except for \dalle (Figure~\ref{fig:APP_color_spatial}C). 

\begin{table}[h]
\footnotesize
\centering
\caption{Chi-squared test results for comparison of accuracy between prompts with no color terms ($Acc_{NC}$, \ie \emph{numeric-simple}) and prompts with color terms ($Acc_C$, \ie \emph{attribute-color}).
N: Number of samples, $\chi^2$: Chi-squared statistic value, Diff?: whether the difference between two accuracies is significant, $\checkmark$: significant difference, =: insignificant difference.}
\include{./tables/appendix/chi_color}\label{tab:app_color}
\vspace{1cm}
\caption{Chi-squared test results for comparison of accuracy between 2-additive prompts with no color terms ($Acc_{2A}$) and 2-additive prompts with color terms ($Acc_{2AC}$).
N: Number of samples, $\chi^2$: Chi-squared statistic value, Diff?: whether the difference between two accuracies is significant, $\checkmark$: significant difference, =: insignificant difference.}
\include{./tables/appendix/chi_2color}\label{tab:app_2color}
\end{table}

\begin{table}[h]
\footnotesize
\centering
\caption{Chi-squared test results for comparison of accuracy between prompts without spatial relationships ($Acc_{NS}$, \ie \emph{2-additive}) and prompts with spatial relationships ($Acc_S$ \ie \emph{attribute-spatial}).
N: Number of samples, $\chi^2$: Chi-squared statistic value, Diff?: whether the difference between two accuracies is significant, $\checkmark$: significant difference, =: insignificant difference.}
\include{./tables/appendix/chi_spatial}\label{tab:app_spatial}
\end{table}
\clearpage

\subsection{Qualitative Analysis of Model Failures}
\label{sec:qual_failures}

\setlength\tabcolsep{1.5pt} 
\begin{table}[h]
\caption{Examples of images where models failed to generate one object as specified in the prompt. See the text for details.}
\begin{tabular}{lccccc}
& A & B & C & D & E \\
1 & \includegraphics[width=.18\linewidth,valign=m]{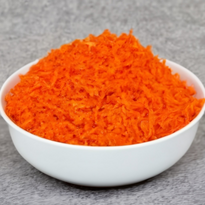} &  
\includegraphics[width=.18\linewidth,valign=m]{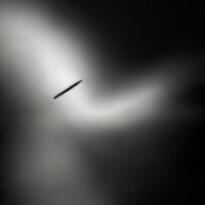} &
\includegraphics[width=.18\linewidth,valign=m]{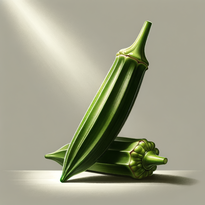} &
\includegraphics[width=.18\linewidth,valign=m]{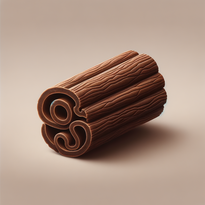} &
\includegraphics[width=.18\linewidth,valign=m]{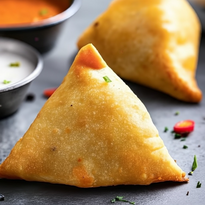} 
\\
2 & \includegraphics[width=.18\linewidth,valign=m]{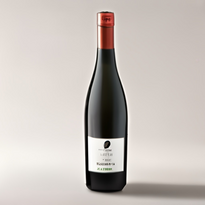} &
\includegraphics[width=.18\linewidth,valign=m]{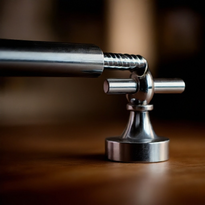} &
\includegraphics[width=.18\linewidth,valign=m]{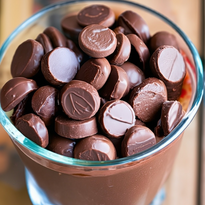} &
\includegraphics[width=.18\linewidth,valign=m]{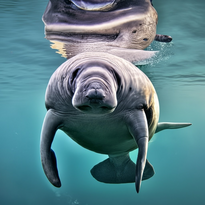} &
\includegraphics[width=.18\linewidth,valign=m]{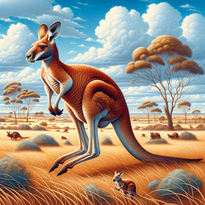} 
\end{tabular}\label{tab:failures}
\end{table}

We manually inspected a subset of images for \emph{numeric-simple} prompts with the ground truth ``1'' (\ie the prompts of the format: \texttt{1 <noun>.}, or \texttt{One <noun>.}) to understand different types of model failures in this most basic case.
One trivial kind of failure, especially observed with some earlier models such as \imagenA and \museA, is not generating the object at all (column A in Table~\ref{tab:failures}).
For example, in A1) the model failed entirely to generate a \emph{spoon}, while in A2) it failed to generate a \emph{corkscrew}. In other instances, the generated object may have some characteristics of the object in the prompt, such as a silhouette of a \emph{pencil} in B1), or the same material, such as metal of a \emph{corkscrew} in B2).

In many other instances, we found that models generate more than one object, such as two \emph{okras} in C1) or many \emph{chocolates} in C2).
Among models we evaluated, Imagen models were most susceptible to generating more than one object when that object was a pistachio, an olive or a cookie.
There were also some more nuanced cases of model failures, subject to interpretation and details of our methodology.
For example, for the \emph{cinnamon stick} in D1), annotators counted 2, 2, 0, 1.5, 1.5, while for the \emph{manatee} in D2) the counts were: 1.5, 1.5, 1, 2, 1.
In both of these examples, rounding up the numbers and taking the most frequent response results in the number ``2'' as the label for that image, although the raw annotations reveal potential ambiguity in the image.

Finally, and as discussed in Appendix~\ref{app:annotation_qualitative}, there were certain ``edge cases'' in counting such as objects that are shown in the background (column E) or shown only partially, where it is possible to count objects in different ways.
For example, in E1), it seems that all annotators assumed that the object in the background is another \emph{samosa}, while in E2) they counted all \emph{kangaroos} that appeared in the image. 
A few cases similar to E2) were observed with \dalle, where the model would generate one dominant entity in the foreground, and several other, much smaller ones in the background.
While our instructions specify that such objects should be counted separately (see Appendix~\ref{app:annotator_instructions} and Figure~\ref{app:count_corner_cases}), but we found that sometimes they were not counted separately or that this was difficult to do  as the boundary between foreground and background may not be apparent in all such images.

We also inspected failure cases in other prompt types. For \emph{additive} prompts we found that \dalle and \museB often correctly depicted two or three  entities as specified in the prompt while failing to depict the correct number of those entities, while some smaller models such as \museA and \imagenA frequently omitted one of the entities entirely. For \emph{attribute-spatial} prompts, which were generally one of the hardest prompts, we notice that \dalle frequently generates a correct (or close to correct) number of entities, but not necessarily in the required spatial arrangement, in contrast to some models that fail to generate those objects at all.
 
\subsection{Technical Resources}
To generate images for all Imagen models and \museA, we used a cluster system containing NVIDIA L4 Tensor Core GPUs.
For \museB, and training experiments in Section~\ref{sec:challenges} we used internal hardware accelerators. We estimate that we used 140 GPU hours to generate all images.

%% file: tables/appendix/chi_representation.tex
\begin{tabular}{lcccccc}
\toprule
 & p-value & $Acc_{D}$ & $Acc_{W}$ & N & $\chi^2$ & Diff? \\
 &  &  &  &  &  &  \\
\midrule
DALLE 3 & 0.827 & 69.8 & 70.4 & 1600 & 0.05 & = \\
\midrule
Midjourney v6 & 0.000 & 64.6 & 78.4 & 1592 & 36.42 & $\checkmark$ \\
\midrule
Imagen-A & 0.001 & 42.2 & 50.4 & 1592 & 10.60 & $\checkmark$ \\
Imagen-B & 0.000 & 35.1 & 45.7 & 1595 & 17.95 & $\checkmark$ \\
Imagen-C & 0.000 & 52.4 & 66.2 & 3178 & 61.85 & $\checkmark$ \\
Imagen-D & 0.000 & 31.4 & 45.1 & 1596 & 31.09 & $\checkmark$ \\
\midrule
Muse-A & 0.018 & 47.4 & 53.4 & 1594 & 5.55 & $\checkmark$ \\
Muse-B & 0.000 & 44.4 & 56.4 & 1600 & 22.56 & $\checkmark$ \\
\midrule
SD1.5 & 0.064 & 33.1 & 37.7 & 1577 & 3.44 & = \\
SD2.1 & 0.000 & 40.9 & 53.4 & 1599 & 24.36 & $\checkmark$ \\
SD 3 & 0.000 & 50.6 & 70.0 & 1600 & 61.92 & $\checkmark$ \\
SDXL & 0.000 & 38.0 & 50.2 & 1564 & 23.09 & $\checkmark$ \\
\bottomrule
\end{tabular}\label{tab:app_has_numeral}

%% file: tables/appendix/chi_freq.tex
\begin{tabular}{lcccccc}
\toprule
 & p-value & $Acc_{R}$ & $Acc_{F}$ & N & $\chi^2$ & Diff? \\
 &  &  &  &  &  &  \\
\midrule
DALLE 3 & 0.189 & 71.7 & 68.6 & 1600 & 1.73 & = \\
\midrule
Midjourney v6 & 0.001 & 67.6 & 75.0 & 1592 & 10.22 & $\checkmark$ \\
\midrule
Imagen-A & 0.000 & 39.4 & 52.5 & 1592 & 26.94 & $\checkmark$ \\
Imagen-B & 0.000 & 32.3 & 47.6 & 1595 & 38.04 & $\checkmark$ \\
Imagen-C & 0.000 & 52.6 & 65.2 & 3178 & 51.90 & $\checkmark$ \\
Imagen-D & 0.621 & 38.9 & 37.6 & 1596 & 0.24 & = \\
\midrule
Muse-A & 0.000 & 41.2 & 58.6 & 1594 & 47.71 & $\checkmark$ \\
Muse-B & 0.000 & 44.9 & 55.4 & 1600 & 17.14 & $\checkmark$ \\
\midrule
SD1.5 & 0.001 & 31.0 & 39.4 & 1577 & 11.72 & $\checkmark$ \\
SD2.1 & 0.000 & 41.5 & 52.3 & 1599 & 18.10 & $\checkmark$ \\
SD3 & 0.011 & 57.0 & 63.3 & 1600 & 6.48 & $\checkmark$ \\
SDXL & 0.373 & 42.9 & 45.2 & 1564 & 0.79 & = \\
\bottomrule
\end{tabular}

%% file: tables/res-additives.tex
\begin{tabular}{l|ccc|ccc|ccc}
\toprule
 &  & SN$\rightarrow$2A & 2A$\rightarrow$3A & SN$\rightarrow$3A \\
 & Number &  &  &  \\
\midrule
\multirow[t]{3}{*}{DALLE 3} & 1 & -9.7 & -3.2 & -12.9 \\
 & 2 & -19.2 & -10.4 & -29.6 \\
 & 3 & -14.2 & -9.5 & -23.7 \\
\cline{1-5}
\multirow[t]{3}{*}{Imagen-C} & 1 & -17.4 & -5.6 & -23.0 \\
 & 2 & -35.4 & -0.1 & -35.5 \\
 & 3 & -21.8 & -9.5 & -31.4 \\
\cline{1-5}
\multirow[t]{3}{*}{Midjourney v6} & 1 & -27.9 & 9.3 & -18.6 \\
 & 2 & -16.7 & -13.0 & -29.7 \\
 & 3 & -17.9 & -10.8 & -28.6 \\
\cline{1-5}
\multirow[t]{3}{*}{Muse-B} & 1 & -0.1 & -10.3 & -10.4 \\
 & 2 & -20.7 & 4.2 & -16.5 \\
 & 3 & -1.6 & -4.1 & -5.7 \\
\cline{1-5}
\multirow[t]{3}{*}{SD3} & 1 & -4.7 & -1.7 & -6.4 \\
 & 2 & -32.8 & -5.0 & -37.8 \\
 & 3 & -18.5 & -6.0 & -24.6 \\
\cline{1-5}
\bottomrule
\end{tabular}

%% file: tables/appendix/chi_color.tex
\begin{tabular}{lcccccc}
\toprule
 & p-value & $Acc_{NC}$ & $Acc_{C}$ & N & $\chi^2$ & Diff? \\
 &  &  &  &  &  &  \\
\midrule
DALLE 3 & 0.470 & 0.77 & 0.79 & 1800 & 0.52 & = \\
\midrule
Midjourney v6 & 0.603 & 0.83 & 0.84 & 1796 & 0.27 & = \\
\midrule
Imagen-A & 0.994 & 0.69 & 0.69 & 1785 & 0.00 & = \\
Imagen-B & 0.499 & 0.59 & 0.62 & 1800 & 0.46 & = \\
Imagen-C & 0.024 & 0.81 & 0.76 & 1999 & 5.07 & $\checkmark$ \\
Imagen-D & 0.717 & 0.55 & 0.54 & 1795 & 0.13 & = \\
\midrule
Muse-A & 0.020 & 0.64 & 0.72 & 1797 & 5.37 & $\checkmark$ \\
Muse-B & 0.298 & 0.73 & 0.69 & 1800 & 1.08 & = \\
\midrule
SD1.5 & 0.048 & 0.59 & 0.52 & 1775 & 3.91 & $\checkmark$ \\
SD2.1 & 0.389 & 0.68 & 0.64 & 1800 & 0.74 & = \\
SD3 & 0.040 & 0.81 & 0.75 & 1794 & 4.20 & $\checkmark$ \\
SDXL & 0.053 & 0.60 & 0.52 & 1787 & 3.73 & = \\
\bottomrule
\end{tabular}

%% file: tables/appendix/chi_2color.tex
\begin{tabular}{lcccccc}
\toprule
 & p-value & $Acc_{2A}$ & $Acc_{2AC}$ & N & $\chi^2$ & Diff? \\
 &  &  &  &  &  &  \\
\midrule
DALLE 3 & 1.000 & 0.48 & 0.48 & 2015 & 0.00 & = \\
\midrule
Midjourney v6 & 0.778 & 0.44 & 0.46 & 2015 & 0.08 & = \\
\midrule
Imagen-A & 0.883 & 0.26 & 0.25 & 2010 & 0.02 & = \\
Imagen-B & 0.036 & 0.42 & 0.31 & 2006 & 4.42 & $\checkmark$ \\
Imagen-C & 0.750 & 0.41 & 0.43 & 2086 & 0.10 & = \\
Imagen-D & 0.023 & 0.51 & 0.38 & 2014 & 5.16 & $\checkmark$ \\
\midrule
Muse-A & 0.212 & 0.53 & 0.45 & 2006 & 1.56 & = \\
Muse-B & 0.061 & 0.67 & 0.56 & 2015 & 3.50 & = \\
\midrule
SD1.5 & 0.144 & 0.25 & 0.18 & 2007 & 2.14 & = \\
SD2.1 & 0.804 & 0.22 & 0.24 & 2007 & 0.06 & = \\
SD3 & 0.225 & 0.42 & 0.50 & 2015 & 1.47 & = \\
SDXL & 0.177 & 0.14 & 0.21 & 2012 & 1.82 & = \\
\bottomrule
\end{tabular}

%% file: tables/appendix/chi_spatial.tex
\begin{tabular}{lcccccc}
\toprule
 & p-value & $Acc_{NS}$ & $Acc_{S}$ & N & $\chi^2$ & Diff? \\
 &  &  &  &  &  &  \\
\midrule
DALLE 3 & 0.095 & 0.40 & 0.37 & 2350 & 2.79 & = \\
\midrule
Midjourney v6 & 0.003 & 0.31 & 0.25 & 2350 & 8.67 & $\checkmark$ \\
\midrule
Imagen-A & 0.000 & 0.19 & 0.07 & 2313 & 74.82 & $\checkmark$ \\
Imagen-B & 0.000 & 0.23 & 0.12 & 2315 & 50.00 & $\checkmark$ \\
Imagen-C & 0.000 & 0.28 & 0.16 & 3084 & 66.50 & $\checkmark$ \\
Imagen-D & 0.000 & 0.31 & 0.17 & 2341 & 56.94 & $\checkmark$ \\
\midrule
Muse-A & 0.000 & 0.28 & 0.17 & 2315 & 37.80 & $\checkmark$ \\
Muse-B & 0.000 & 0.37 & 0.27 & 2350 & 27.00 & $\checkmark$ \\
\midrule
SD1.5 & 0.003 & 0.13 & 0.09 & 2325 & 8.94 & $\checkmark$ \\
SD2.1 & 0.006 & 0.13 & 0.09 & 2346 & 7.63 & $\checkmark$ \\
SD3 & 0.003 & 0.31 & 0.26 & 2348 & 8.74 & $\checkmark$ \\
SDXL & 0.000 & 0.15 & 0.10 & 2328 & 13.87 & $\checkmark$ \\
\bottomrule
\end{tabular}

%% file: app_human_annotation.tex
\section{Collecting Human Annotations}
\label{app:annotations}

\subsection{Annotation Instructions}
\label{app:annotator_instructions}

\begin{figure}[p]
\centering
\includegraphics[width=.75\linewidth]{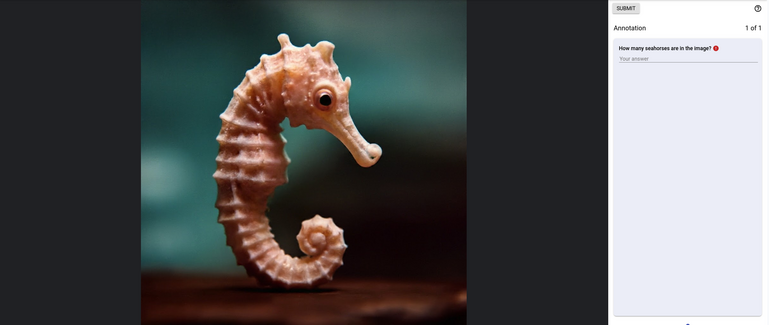}
\caption{A screenshot of the user interface for one annotation example for Annotation Task 1. The response to the question is given in a free-form text format.}\label{fig:screenshot1}
    \par
    \vspace{1cm}
\includegraphics[width=.75\linewidth]{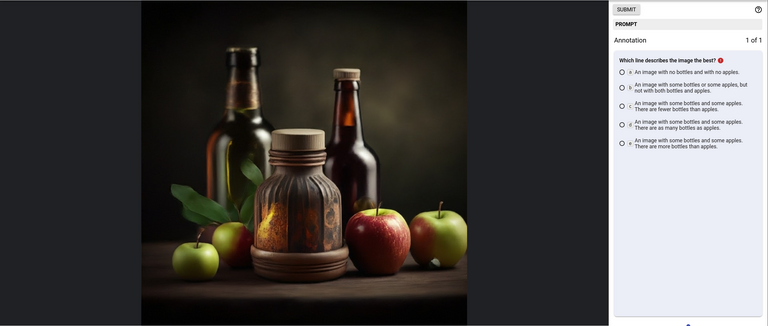}
\caption{A screenshot of the user interface for one annotation example for Annotation Task 2. The response to the question is a single radio-button choice.}\label{fig:screenshot2}
    \par
    \vspace{1cm}
\includegraphics[width=.75\linewidth]{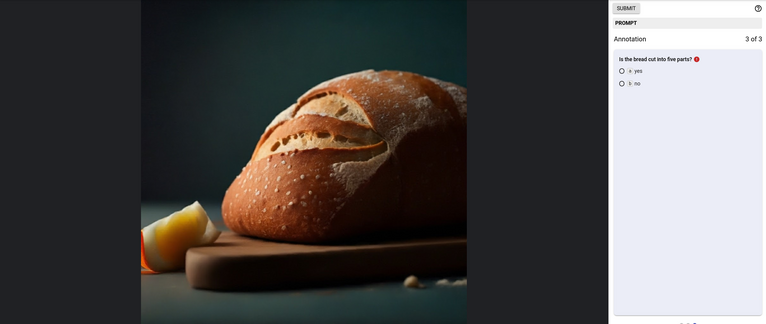}
\caption{A screenshot of the user interface for one annotation example for Annotation Task 3. The response to the question is a single radio-button choice with yes/no options. In this example, there were three questions for this image, and this was the third question.}\label{fig:screenshot3}
\end{figure}

All human annotators received training before completing the annotation tasks. 
They were instructed on how to use the web interface, were given task-specific instructions and were shown several examples of expected annotations with explanations.
After the training, we conducted a pilot study with a small subset of images to confirm that task instructions were understood and followed.
All images shown to annotators were passed through safety filters, and flagged images were manually inspected to ensure that no offensive, harmful or otherwise problematic content would be present in images.
We did not collect any personally identifiable information.

\begin{figure}[h]
    \centering
    \includegraphics[width=.8\linewidth]{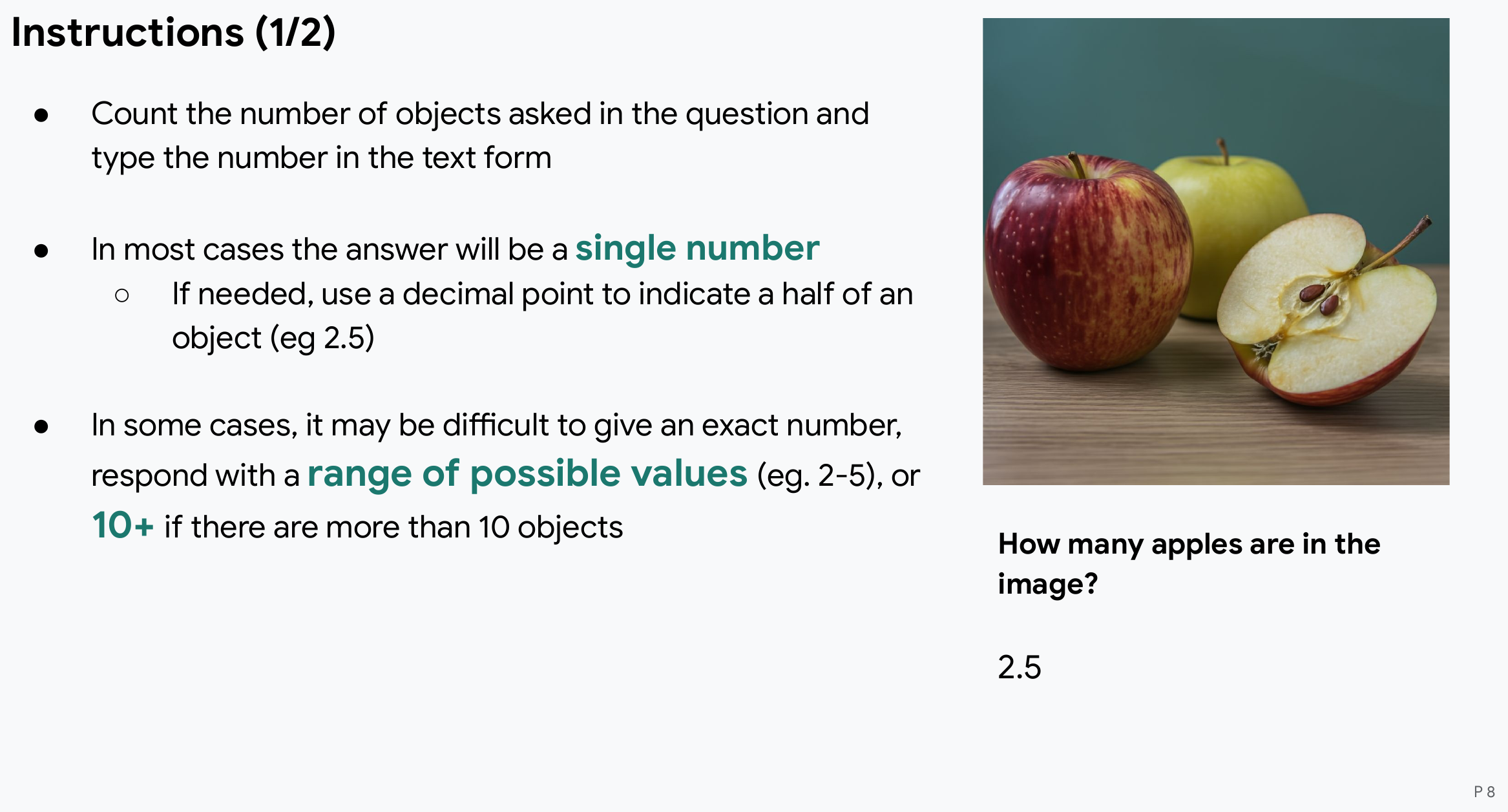}
    \par
    \vspace{1cm}
    \includegraphics[width=.8\linewidth]{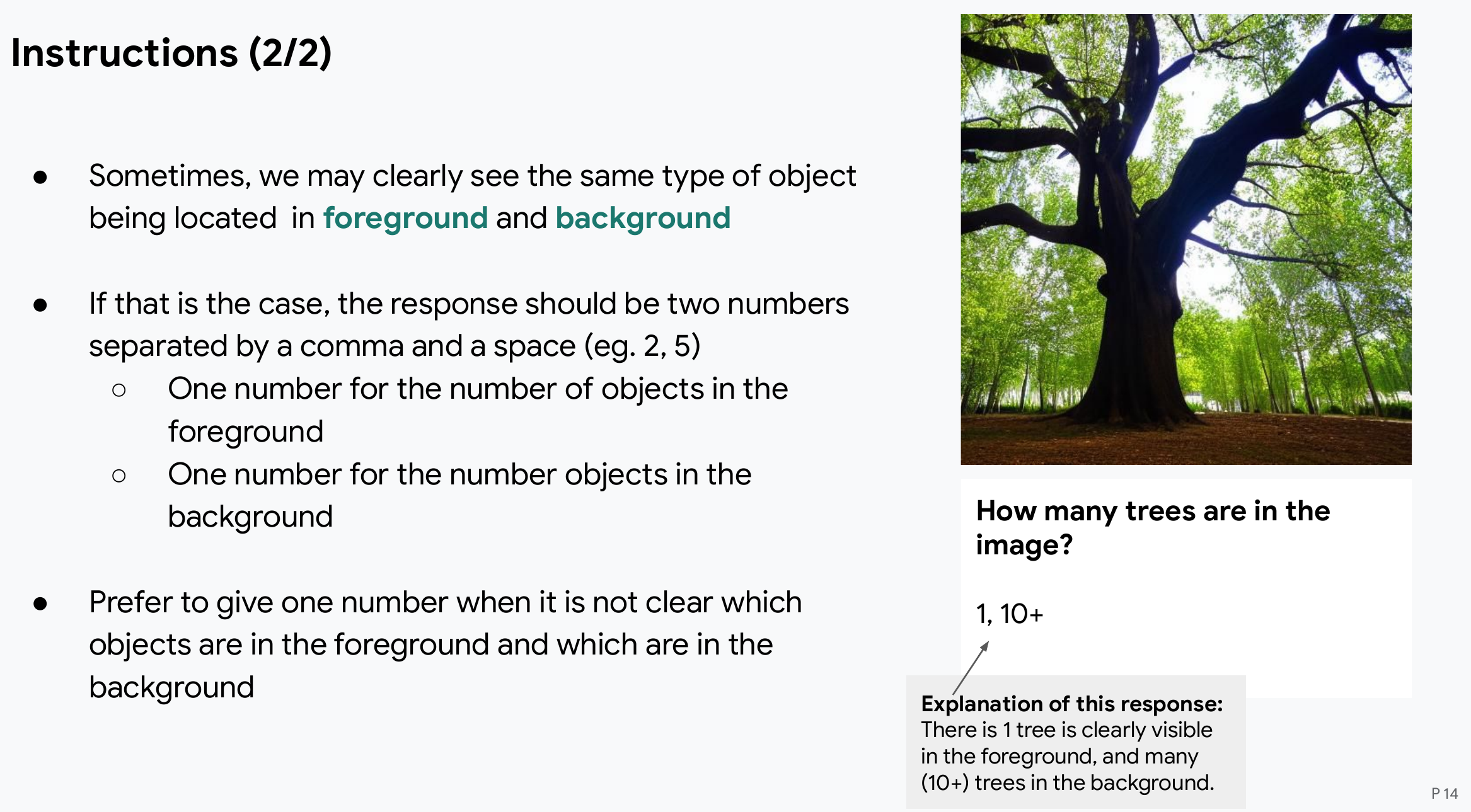}
    \par
    \vspace{1cm}
    \includegraphics[width=.8\linewidth]{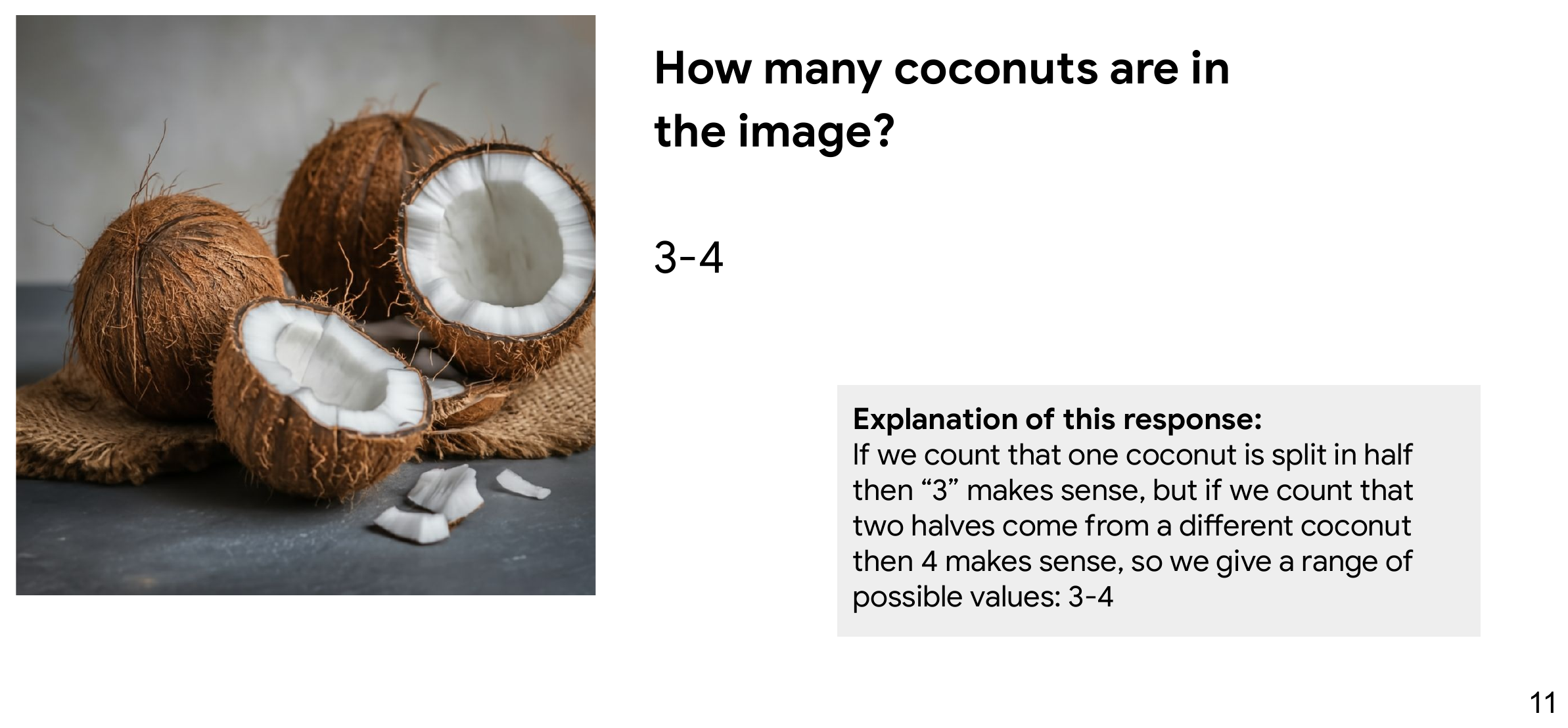}
    \par
    \vspace{1cm}
    \caption{Annotation instructions including examples of edge cases in counting in Task 1.}
    \label{app:count_corner_cases} 
\end{figure}

\paragraph{Annotation Task 1 Instructions.}
Annotation instructions, including a few examples of edge cases in counting are shown in Figure~\ref{app:count_corner_cases}.
The goal of including a few examples of edge cases was to establish a  guideline on how to count objects in cases where several interpretations are possible.
For example, if it is possible that two shown halves of an object come from the same object we instructed annotators to count it as one object.
If it is unlikely that they come from the same object, we instructed them to count the two parts as two different objects.
Here, our aim was to provide guidance to reasonably standardize responses in such cases, but we did not aim to exhaustively cover all possible cases as we also wanted annotations to reflect individual differences in object perception in such less precise or ambiguous cases.
If there were more than 10 objects present in the image, the instruction was to respond with ``10+''. In this task, responses were provided as free-form text.
An example screenshot from this task is shown in  Figure~\ref{fig:screenshot1}.

\paragraph{Annotation Task 2 Instructions.}
\begin{figure}[h]
    \centering
    \includegraphics[width=.8\linewidth]{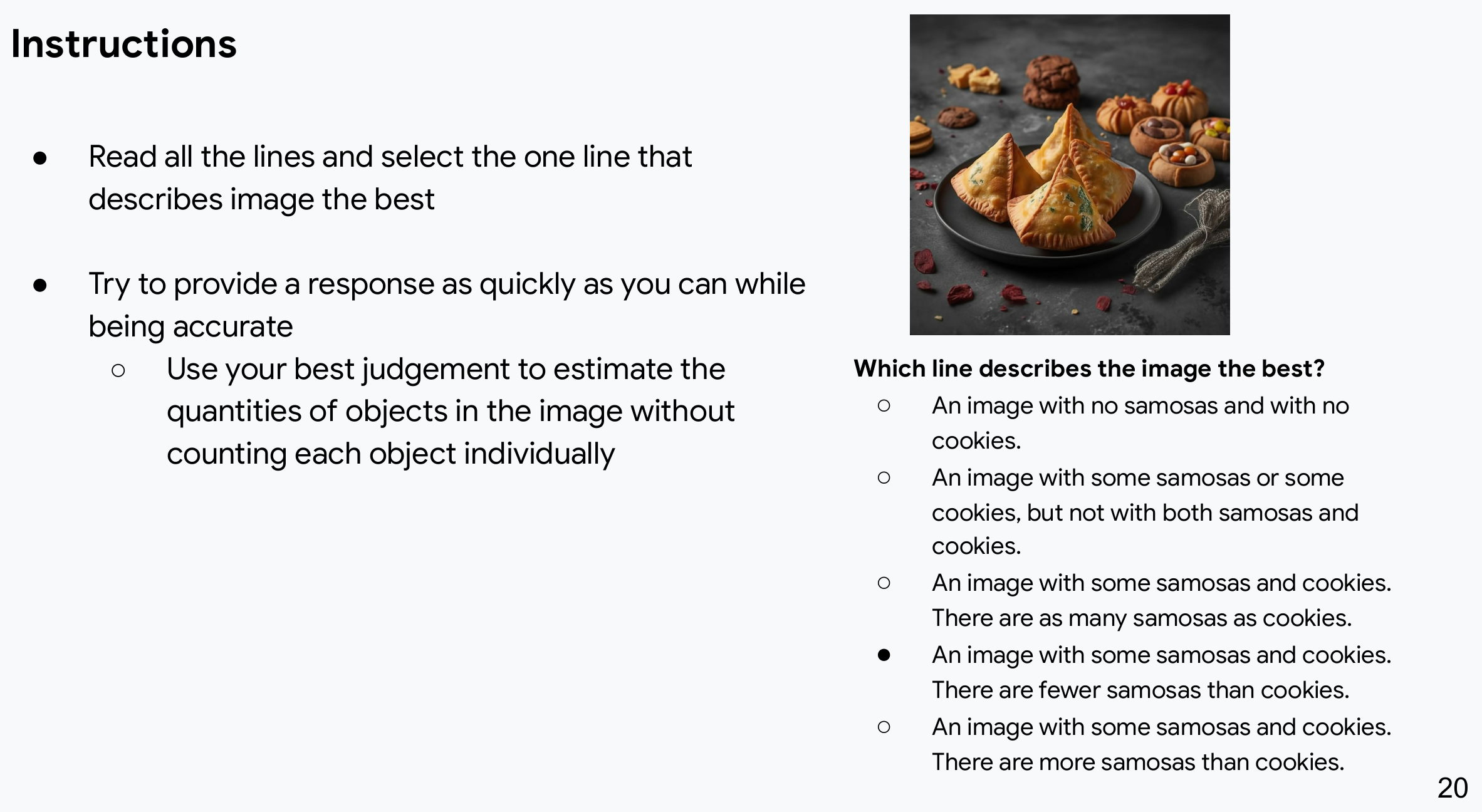}
    \par
    \vspace{1cm}
    \includegraphics[width=.8\linewidth]{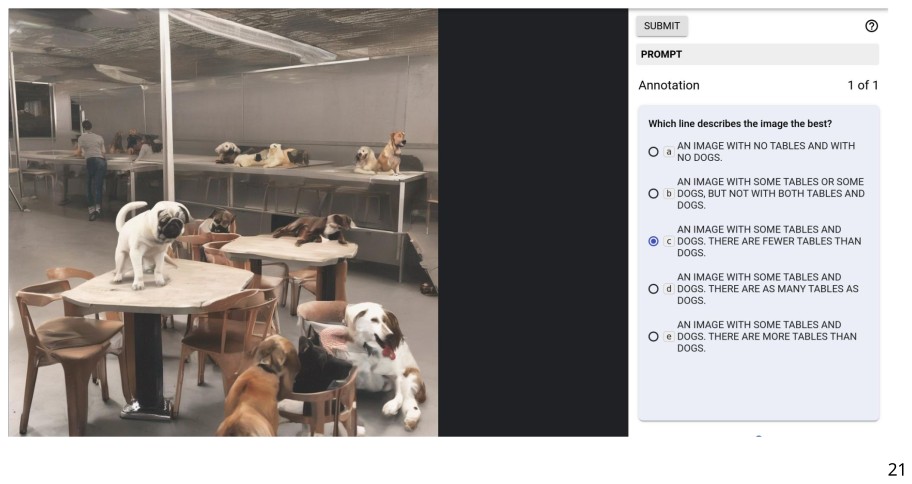}
    \caption{Annotation instructions including an example Task 2.}    
    \label{app:estimation_instruction} 
\end{figure}

Instructions shown to annotators in the second task are shown in Figure~\ref{app:estimation_instruction}. Annotators were told to provide a response quickly while being accurate, to encourage them to estimate quantities on this task, as opposed to counting each object in the image separately.
We wanted to get a response that best reflects judgement of approximate quantities, as opposed to exact quantities that were the subject of interest in Task 1. An example screenshot from this task is shown in  Figure~\ref{fig:screenshot2}.

\paragraph{Annotation Task 3 Instructions.}
\begin{figure}[h]
    \centering
    \includegraphics[width=.95\linewidth]{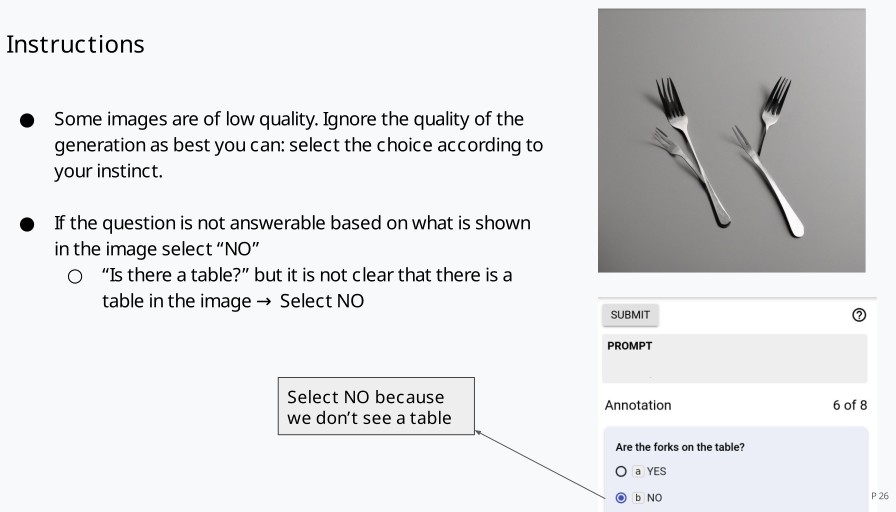}
    \par
    \vspace{1cm}
    \includegraphics[width=.95\linewidth]{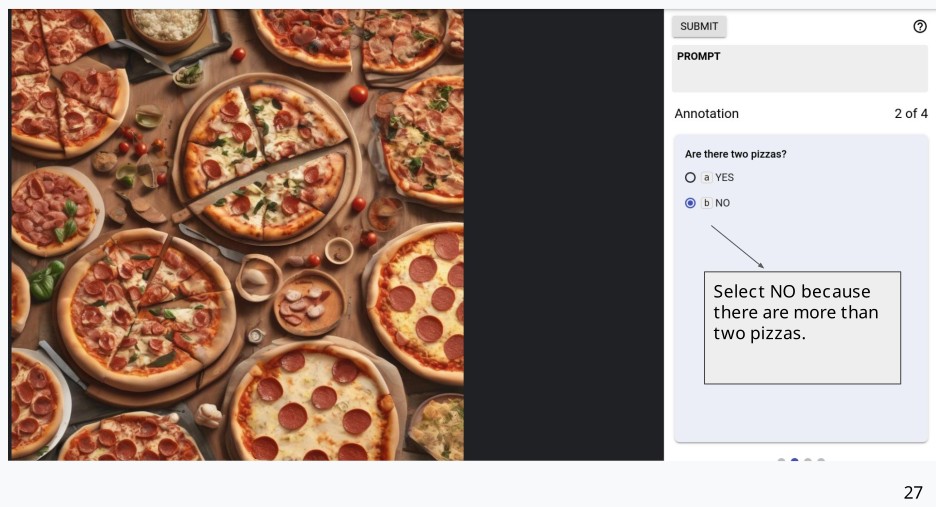}
    \caption{Annotation instructions including examples in Task 3.}    
    \label{app:consistency_instruction} 
\end{figure}
Figure~\ref{app:consistency_instruction} shows instructions and an example annotation in Task 3, as well as an additional example screenshot in Figure~\ref{fig:screenshot3}.

\clearpage
\subsection{Quantitative Analysis}
\label{app:data_analysis}

\begin{minipage}{\textwidth}
\begin{minipage}[b]{0.48\textwidth}
\centering
\includegraphics[width=\textwidth]{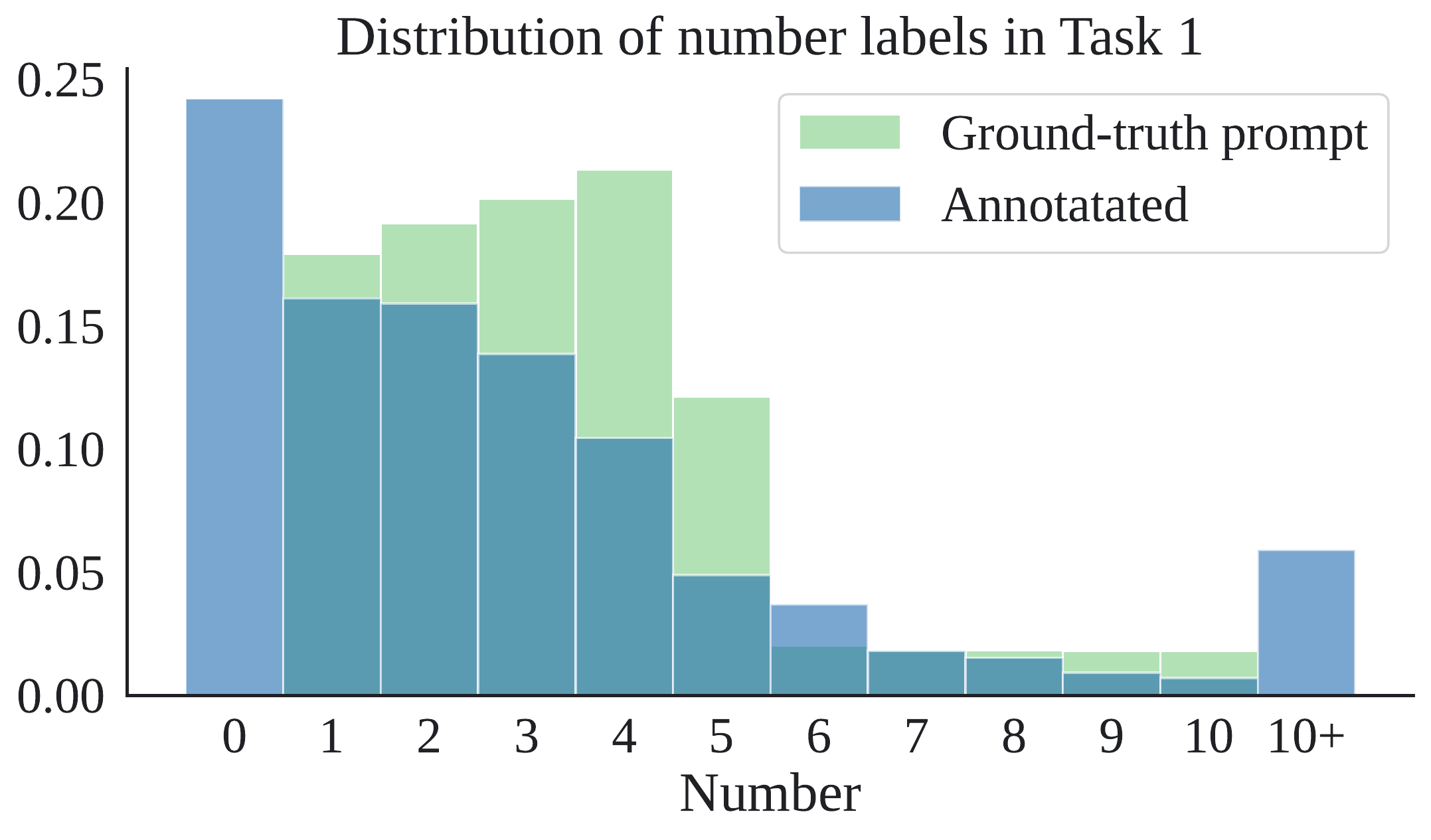}
\captionof{figure}{Distribution of numbers in ground-truth prompts used to generate images, and numbers given as counts of objects in annotated images.}\label{fig:app_annotations}
\end{minipage}
\hfill
\begin{minipage}[b]{0.5\textwidth}
    \centering
    \captionof{table}{Percentage of annotations where n out of 5 annotators provided the same response in Annotation Task 1.}\label{tab:app_annotations}
    \include{./tables/appendix/task1_agreement}
\end{minipage}
\end{minipage}

In total, in Annotation Task 1 we collected \num{718729} annotations, ranging from approximately $56-58$K annotations per model, except for \imagenC where we collected $91$K annotations as five additional images per prompt were generated for evaluation and fine-tuning experiments in Section~\ref{sec:challenges} (for a total of 10 seeds for \imagenC). 
In Annotations Tasks 2 and 3, we collected \num{20960} and \num{62010} annotations, respectively.
For \imagenC, the analyses and statistics below will be based on using five seeds for each model, although the data we release also includes additional images and annotations.
Released data will not contain any images showing faces (approximately 122 images), due to potential privacy concerns, but analyses below were done including annotations collected for those images.

\paragraph{Response processing.} From raw responses we removed spaces and all non-numeric characters, except for the letters ``o''/``O'' which we replaced with the number 0, ``10+'' entries were replaced 11 to get a numerical format, and ranges (\ie 2--4) were converted into an integer by rounding up the average of the two values.
When two comma-separated numbers were given, we took only the first number. If after these steps the response was an empty string, we removed it from the dataset. In total, only 10 such responses were removed.

\paragraph{Response aggregation.} 
Each image and each question was shown to five annotators, collecting five annotations in total per image--question pair. In Task 1, \emph{additive} prompts had several questions, one for each entity in the image, while in Task 3 there were several questions derived from the original text prompt.
To get a single number representing the number of objects in an image we used the mode of five numbers in Tasks 1 and 2. 
In Task 2 those numbers were numerically encoded single-choice radio button options.
In total, after aggregating responses there were \num{143748} labels in Task 1, and \num{4192} labels in Task 2.
In Task 3, we averaged all responses across all questions to get a score and refer to it as ``accuracy'' for consistency in analyses.
Figure~\ref{fig:app_annotations} shows normalized distributions of numbers in the original text prompts (green) and the distribution of numbers in annotated data (blue).

\paragraph{Annotator agreement.} We observe high level of agreement among annotators. In Task 1, for each image--question pair, all 5/5 raters gave the same response in 77.5\% of cases, at least 4/5 raters gave the same response in 87.9\% of cases and at least 3/5 in 96.5\%. In Task 2, the analogous percentages were as follows: 74.7\%, 89.5\% and 99.3\%, and in Task 3 where we only had binary responses all 5/5 raters agreed in 84.9\% of cases, while at least 4/5 agreed in 94.9\%.
Expressed as Krippendorff's alpha coefficient of inter-annotator agreement, $\alpha=0.865$ for Task1, $\alpha=0.863$ for Task 2, and $\alpha=0.903$ for Task 3.
A breakdown per model annotator agreements in Task 1 is shown in Table~\ref{tab:app_annotations}.

\include{./tables/appendix/qual_task1}
\include{./tables/appendix/qual_task2}
\include{./tables/appendix/qual_task3}

\subsection{Qualitative Analysis}
\label{app:annotation_qualitative}

While the overall level of disagreement among annotators was low, as shown in Table~\ref{tab:app_annotations}, we manually inspected annotations to understand potential reasons for annotator disagreement and ensure the overall quality of the annotations. 
Examples for each task are shown in Figures~\ref{fig:app_exampes_task1}--\ref{fig:app_exampes_task3}, and below we discuss specific cases and the overall observed trends.

\subsubsection{Task 1: Exact Number Generation}
\label{app:task1_qual}
Figure~\ref{fig:app_exampes_task1} shows 10 selected examples of annotations with a varying level of disagreement in Task 1.
We selected these examples as a representative subset of images, and discuss some common reasons why we believe there was some level of disagreement related to the perception of objects or quantities in an image.
Some of these observations, such as that most people only count objects when those objects are clearly identifiable in an image, also held for the analysis of disagreement in Tasks 2 and 3.

First, we find that some disagreement may arise from a different interpretation of instructions, such as two annotators counting one half of a coconut as a whole, while three annotators counting it as a half, as per instructions (Example 1 in Table~\ref{fig:app_exampes_task1}).
In other cases, where there is some ambiguity regarding the identity of objects, it appears that annotators only counted objects when they were certain that shown objects matched the identity of objects mentioned in the question (Example 2 in the table).
There were images where it was difficult to count objects based on how salient such objects were in an image, such as the trees shown in Example 3. We attempted to account for some such cases in instructions (such as dividing objects between those in the foreground and those in the background), but even then the boundary between foreground and background was not always apparent in generated images.
Sometimes objects appeared distorted (Example 4), or there were morphed objects in an image and it was not possible to distinguish between individual entities (Examples 5 and 6).

There was a noticeable discrepancy in labels for images containing eggs, such as in Example 7.
Specifically, we noticed that all annotators counted eggs when those appeared in shells or they were boiled, but if eggs were cracked (in the form of an egg whites and yolk) or fried, they would frequently not be counted as eggs. 
When we asked our annotators about the reasons for counting eggs in this way, we were told that some annotators perceived fried eggs as a dish, distinct from the concept of a ``raw egg''.

We also found that some level of disagreement stems from individual differences in color perception, such as ``white apples'' in Example 8.
Finally, we also found that there were rare instances of incorrect annotations. For example in Example 9 we expect the correct answer to be ``4'' for the third question ``How many chocolates are to the left of the fish?'', but only one response matched our expectation.
As well, Example 10 shows another instance where there is some ambiguity in interpretation, as the expected answer here is ``1'' and only one response matched the expectation.

\subsubsection{Task 2: Approximate Number Generation and Zero}
\label{app:task2_qual}
Figure~\ref{fig:app_exampes_task2} shows six examples of selected images and corresponding annotations where there was some level of disagreement among annotators.
We highlight these examples as representative of the types of disagreements we see in the data.
When objects in images are not clearly identifiable people tend to indicate that one of the two objects is not present, such as Examples 1--3 in the figure.

In Examples 1 and 2, the model generated objects that were morphed forms of the two entities in the prompt. In Example 1, many of the objects appear to be approximately shaped like trowels, with one in the lower left corner being shaped like a manatee, but all objects have the color of a manatee.
In this example, 3/5 people indicated that they do not recognize manatees, trowels or either.
Somewhat similar trend is observed in Example 2, where generated objects appear to be a morphed combination of the two entities mentioned in the prompt--fish and seahorses.
However, in this image more people were able to recognize objects, though from the data we have it is unclear how ambiguous objects were perceived.

In Example 3 we see another instance where the majority of annotators did not perceive an object when that object was not clearly identifiable.
Specifically we see guard rails, but it is not clear that those rails belong to a crib, and so only one annotator indicated that there are more kangaroos, while the remaining four indicated that there are ``no kangaroos or no cribs``.

In Example 4, we see some individual differences in perception of quantity, where some people perceive the quantity of flowers as ``many'' while other as ``only a few''.
In Example 5, and as in images in Task 1, we again observe that 4/5 annotators did not consider ``fried eggs'' as eggs, with the exception of one annotator.
Finally, Example 6 is interesting because it highlights that people naturally expect to see watermelon seeds in a watermelon, even though the objects in the watermelon do appear as some kind of seeds.

\begin{figure}
    \centering
    \begin{minipage}{0.5\textwidth}
        \centering
        \includegraphics[width=\textwidth]{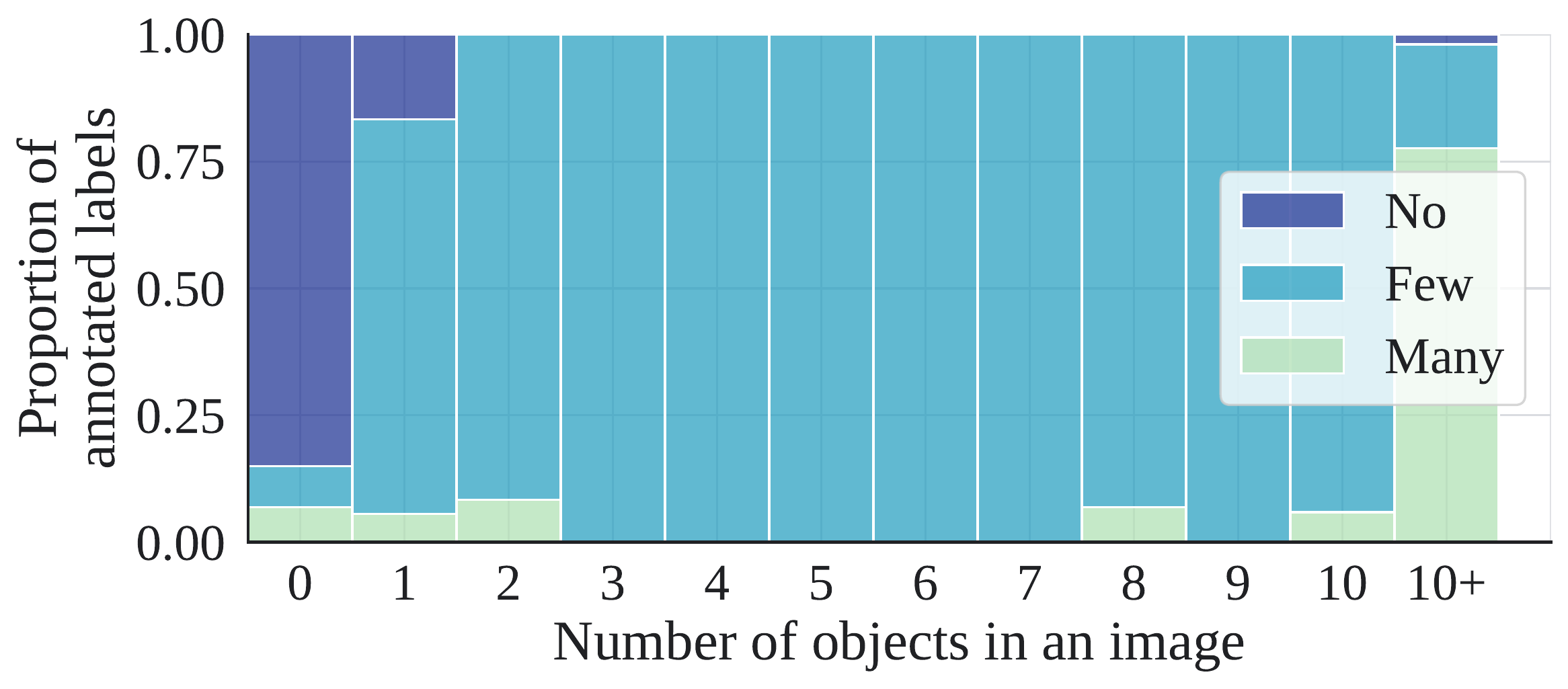}
\caption{The proportion of ``no'', ``few'' and ``many'' annotated labels associated with a specific count of objects in an image.}\label{fig:app_task2_counts_humans}
    \end{minipage}\hfill
    \begin{minipage}{0.47\textwidth}
        \centering
        \vspace{.2cm}
        \includegraphics[width=\textwidth]{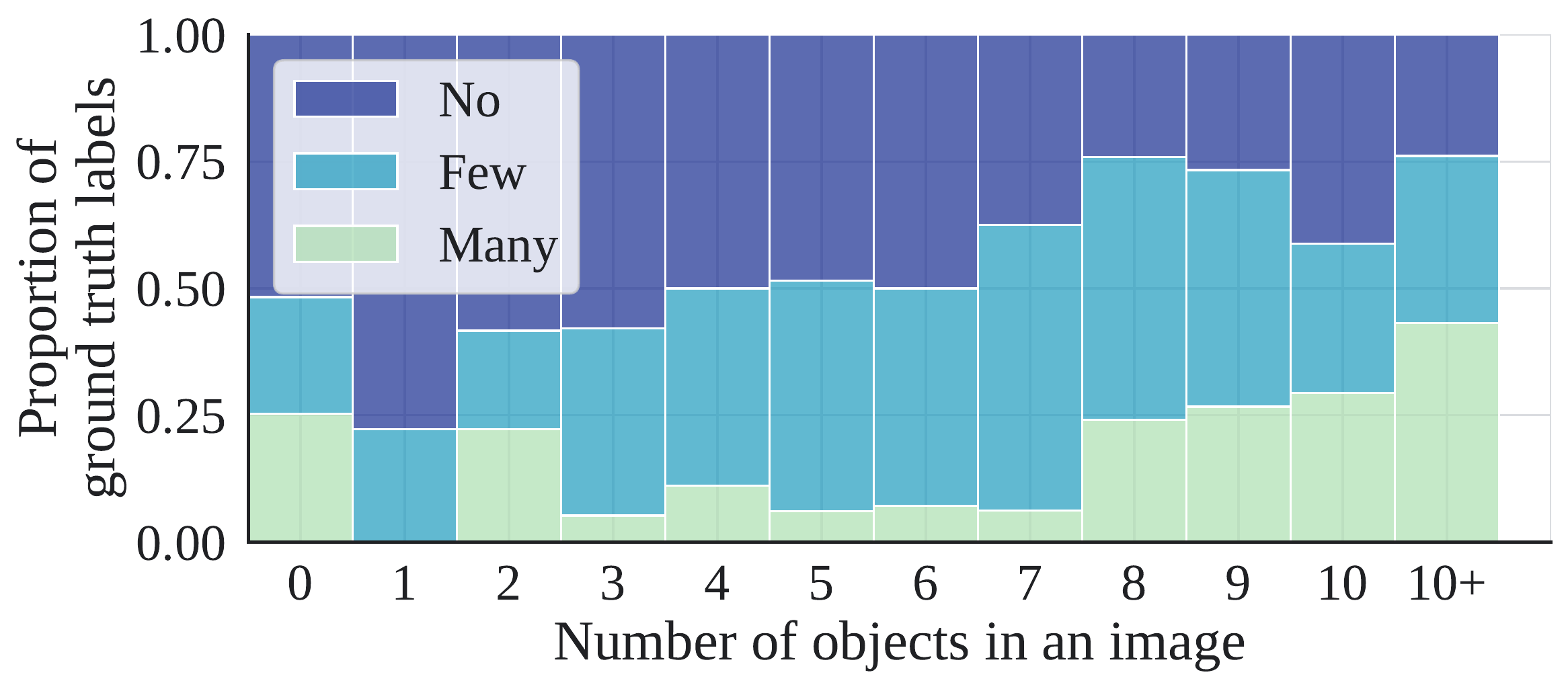} 
        \caption{The proportion of ``no'', ``few'' and ``many'' labels from the ground truth prompts associated with a specific count of objects in an image.}\label{fig:app_task2_counts_models}
    \end{minipage}
\end{figure}

As in this task we used linguistic quantifiers to study approximate quantities, there is some inherent subjectivity involved when interpreting the quantities.
Specifically, as seen in Example 4, some people perceive that there are ``many'' flowers in the vase, while others perceive ``only a few'' flowers.

To further understand these individual differences, we collected additional annotations for images generated for \emph{approx-1-entity} prompts  in Task 2.
These are the prompts that only contain one entity and the quantity associated with the entity is either ``no'', ``[only a] few'' or ``many''.
Additional annotations were counts of objects in those images, as done in Task 1.
Then, we paired annotations collected in Task 2, namely labels such as ``no'', ``few'' and ``many'' extracted from the lines annotators selected when completing the task with those counts, and plot the proportion of labels for each count in Figure~\ref{fig:app_task2_counts_humans}.\footnote{For each image we had five annotations, but we take the most frequent annotation as the label.}

Despite differences in individual interpretations of linguistic quantifiers, we find that annotators were highly consistent in their interpretations of such quantities--the overwhelming majority of them selected ``no'' when they counted zero objects in an image (the dark blue bar over ``0'' in Fig.~\ref{fig:app_task2_counts_humans}), and similarly, when there are more than 10 objects annotators labelled those images as ``many''.
The notion of ``few'' had a more distributed span ranging from the numbers 1 to 10, but for counts in the range 3--7 100\% of labels were ``few''. 

We performed a similar analysis for labels extracted from the ground truth prompts, instead of annotations, for a comparison on how generated quantities in images correspond to labels.
The results are shown in Figure~\ref{fig:app_task2_counts_models}.
We indeed see a trend where the bars for ``no'' are more skewed towards the left, meaning more models generated fewer items when the word ``no'' was in the prompt, and analogously  we see more ``many'' items toward the right.

\subsubsection{Task 3: Conceptual Quantitative Reasoning}
\label{app:qual_task3}

Figure~\ref{fig:app_exampes_task3} shows six examples of images and corresponding annotations in Task 3.
We generally see some ambiguity in responses when images are supposed to depict broken objects.
For instance, in Example 1, 2/5 annotators responded that there is one plate in the image, while the other three annotators responded that there are pieces of a plate and that the plate was broken.
Similarly in Example 2, 2/5 annotators perceive that a pencil is broken when pencil lead was cracked. 

Some disagreement was present when talking about parts, as seen in Examples 3--5.
While the loaf of bread in Example 3 is cut into three parts, the parts are not equal, but 2/5 annotators responded that the loaf of bread is cut into thirds.
Examples 4 and 5 highlight the difficulty of model evaluation in this task.
In Example 4, the question where annotators gave different responses (``Are the pieces inside the apple?'') is neither meaningful nor relevant for evaluation of numerical reasoning, and some level of disagreement in this case might be expected.

In Example 5, we speculate that there were individual differences in perception of ``half of a certain color''---2/5 annotators did indicate that one half of the pencil is red and the other half is blue, while 3/5 did not. We manually inspected images and annotations for this specific prompt to see what would an image look like if all annotators agreed. One such example is shown in the last row of the table. We speculate that in answering this question, the annotators might have considered the color of pencil lead as well as the wood when answering the question.

Based on these and other similar examples, we conclude that disagreement in this task may come from ambiguity present in images, questions or both.
The ambiguity in images may arise because objects and their parts are not clearly shown, as observed in Tasks 1 and 2, but also because it may not be clear whether objects are broken/sliced etc. The ambiguity in questions may arise because questions are not sufficiently specific, or in some cases, not meaningful.

\subsection{Methodological Challenges in Model Evaluation}
\label{app:model_eval}

In Tasks 1 and 2 we could directly express model accuracy by comparing whether the annotated count (\eg ``3'') or quantity (\eg ``many'') matched the corresponding count or quantity in the ground truth prompt used to generate the image.
This was more difficult in Task 3, where prompts included objects that were shown in parts or pieces, and these parts or pieces might be associated with different properties, such as different sizes or colors.
Text-to-image models which generate images that better depict more of such properties should be given a higher score, compared to those that match only some.
For example, let us consider the following text prompt: \mbox{\emph{A pizza cut into 3 slices.}}%

Given this prompt, if a model A produces an image of a pizza that is not sliced at all, and a model B produces an image with a pizza cut into quarters, we would expect that model B is better aligned with the prompt as it correctly captured the notion of ``cut'' and ``slice''.
For this reason, we decided to ask several questions that are grounded in the text prompt for each image in evaluation of models on this task.
To obtain such questions, we use a recent automatic method based on the Davidsonian Scene Graph (DSG) ~\cite{cho2023davidsonian}.
This method generates questions based on words in the prompt in such a way that the expected answer to the question is ``yes''.
For the example prompt above, such questions could be: \emph{Is there a pizza?}, \emph{Is the pizza cut?} and \emph{Is the pizza cut into 3 slices?}
Model B would score higher as it generated an image that can be answered with ``yes'' in 2/3 questions, while for model A this would only be the case for 1/3 questions.

While the majority of automatically generated questions are relevant in the context of evaluation of numerical reasoning, there are some limitations to this approach, as hinted in Section~\ref{sec:results_task3} and in Appendix~\ref{app:qual_task3}.
First, some questions might be less discriminative for more capable text-to-image models that are able to clearly depict the object in the prompt. Specifically, it may be the case that for all models the answer to the question \emph{Is there a pizza?} will be ``yes''.
Second, occasionally the answer to some questions might be ``no'', as seen in Examples 5 and 6 in Table~\ref{fig:app_exampes_task3} where questions imply that the pencil might be split into halves (\ie \emph{Is there half of a pencil?} and \emph{Is the pencil in two halves?}).
We also noticed that some questions were not informative or were confusing, such as \emph{Are the pieces inside the apple?} (Example 4 in Table~\ref{fig:app_exampes_task3}) or \emph{Is the quarter of a loaf of bread on the loaf of bread?}
Based on manual inspection of the data, such questions accounted for a small proportion of all questions and we expect that as automatic methods advance their number will be even smaller in the future.

%% file: tables/appendix/task1_agreement.tex
\begin{tabular}{lrrrrr}
\toprule
 & 1/5 & 2/5 & 3/5 & 4/5 & 5/5 \\
 &  &  &  &  &  \\
\midrule
DALLE 3 & 0.0 & 2.3 & 6.5 & 8.5 & 82.7 \\
\midrule
Midjourney v6 & 0.1 & 2.4 & 6.6 & 9.3 & 81.7 \\
\midrule
Imagen-A & 0.1 & 2.3 & 6.6 & 8.7 & 82.3 \\
Imagen-B & 0.1 & 2.7 & 7.4 & 9.4 & 80.4 \\
Imagen-C & 0.1 & 2.8 & 8.1 & 10.4 & 78.7 \\
Imagen-D & 0.1 & 3.8 & 7.6 & 9.4 & 79.1 \\
\midrule
Muse-A & 0.2 & 3.2 & 8.6 & 11.0 & 77.0 \\
Muse-B & 0.1 & 3.3 & 10.1 & 12.0 & 74.5 \\
\midrule
SD1.5 & 0.2 & 4.2 & 10.1 & 11.5 & 73.9 \\
SD2.1 & 0.1 & 4.5 & 11.8 & 12.6 & 71.0 \\
SDXL & 0.3 & 5.1 & 9.7 & 11.0 & 73.9 \\
SD3 & 0.2 & 3.8 & 9.7 & 11.3 & 75.0 \\
\bottomrule
\end{tabular}

%% file: tables/appendix/qual_task1.tex
\begin{figure}[h]
\renewcommand{\arraystretch}{2}
\centering
\resizebox{\textwidth}{!}{
\begin{tabular}{ccccc}
\toprule
Example \# & Prompt & Image & Questions & Answers \\ \bottomrule
 &  & \multirow{3}{*}{\includegraphics[width=0.15\textwidth]{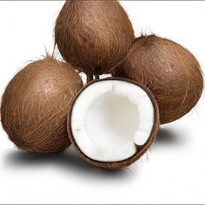}} &  &   \\
1  & 10 coconuts. &  & How many coconuts are in the image?  & 3.5, 3.5, 4, 4, 3.5 \\
 &  &                        &     &    \\ \midrule
  &  & \multirow{3}{*}{\includegraphics[width=0.15\textwidth]{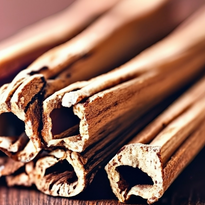}} &  &   \\
2  & 5 cinnamon sticks. &  & How many cinnamon sticks are in the image?  & 0, 0, 5.5, 0, 6 \\
 &  &                        &     &    \\ \midrule
  &  & \multirow{3}{*}{\includegraphics[width=0.15\textwidth]{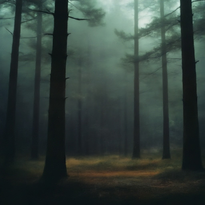}} &  &   \\
3  & 5 trees. &  & How many trees are in the image?  & 7.5, 7.5, 8, 8.5, 11 \\
 &  &                        &     &    \\ \midrule
&  & \multirow{3}{*}{\includegraphics[width=0.15\textwidth]{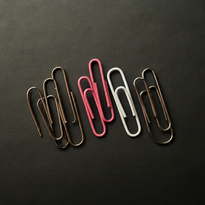}} &  &   \\
4  & 8 paperclips. &  & How many paperclips are in the image?  & 8, 6.5, 9, 7, 6 \\
 &  &                        &     &    \\ \midrule
 &  & \multirow{3}{*}{\includegraphics[width=0.15\textwidth]{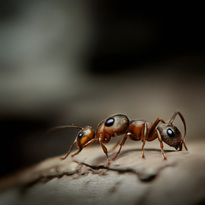}} &  &   \\
5  & Two ants. &  & How many ants are in the image?  & 1.5, 2, 1.5, 2, 2.5 \\
 &  &                        &     &    \\ \midrule
  &  & \multirow{3}{*}{\includegraphics[width=0.15\textwidth]{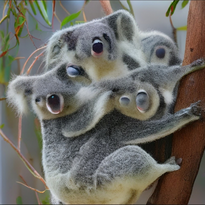}} &  &   \\
6  & 9 koalas. &  & How many koalas are in the image?  & 3.5, 4, 3.5, 2.5, 4 \\
 &  &                        &     &    \\ \midrule
 &  & \multirow{3}{*}{\includegraphics[width=0.15\textwidth]{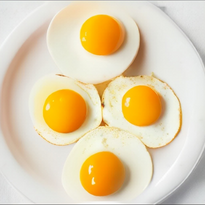}} & &  \\
7  &  8 eggs. & & How many eggs are in the image?  & 4, 0, 4, 4, 0 \\
 &  & &     &    \\ \midrule
  &  & \multirow{3}{*}{\includegraphics[width=0.15\textwidth]{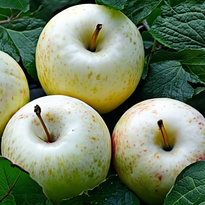}} &  How many apples are in the image?  & 3.5, 4, 3.5, 4, 3-4   \\
8  & Four white apples. &   & How many white apples are in the image?    & 3.5, 0, 0, 0, 3.5 \\
 &  &       &    &  \\ \midrule
   &  & \multirow{4}{*}{\includegraphics[width=0.15\textwidth]{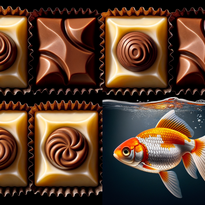}} & How many chocolates are in the image?  &  6, 6, 6, 6, 6 \\
   & There are 4 chocolates  &   & How many fish are in the image?  & 1, 1, 1, 1, 1  \\
 9  & to the left of one fish. &   & How many chocolates are to the left of fish?	  & 4, 0, 0, 2, 0  \\
 &  &       &  How many fish are to the right of chocolates?  &  1, 1, 1, 1, 1\\ \midrule
   &  & \multirow{3}{*}{\includegraphics[width=0.15\textwidth]{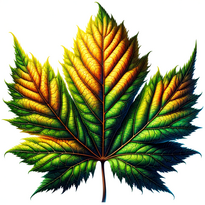}} &  &   \\
 10  & 1 leaf. &   & How many leaves are in the image?  & 7, 7, 7, 7, 1  \\
 &  &       &    &  \\
\bottomrule
\end{tabular}}
\caption{Examples of images and annotations in Annotation Task 1. Only images and questions were shown to annotators when completing the task.}
\label{fig:app_exampes_task1}
\end{figure}

%% file: tables/appendix/qual_task2.tex
\begin{figure}[h]
\renewcommand{\arraystretch}{1.6}
\centering
\resizebox{\linewidth}{!}{
\begin{tabular}{cccl}
\toprule
Example \# & Prompt & Image & Answer (Best line match) \\
\bottomrule
 & & \multirow{5}{*}{\includegraphics[width=0.2\textwidth]{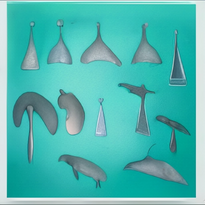}} &   
some trowels or some manatees, but not with both\\
  &  &  & 
no trowels and no manatees \\
1 & \makecell{An image with some trowels and some manatees.\\There are as many trowels as manatees.
}  &  &  
as many trowels as manatees \\
 & & & 
some trowels or some manatees, but not with both\\
 & & &
more trowels than manatees \\
 \midrule
 & & \multirow{5}{*}{\includegraphics[width=0.2\textwidth]{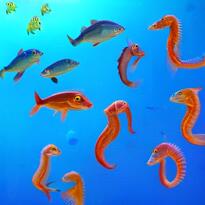}} &   
 more fish than seahorses
 \\
  &  &  & 
 fewer fish than seahorses\\
2 & \makecell{An image with some fish and some seahorses.\\There are more fish than seahorses.}  &   &  
as many fish as seahorses\\
 & & & 
some fish or some seahorses, but not with both
 \\
 & & &
as many fish as seahorses\\
 \midrule
& & \multirow{5}{*}{\includegraphics[width=0.2\textwidth]{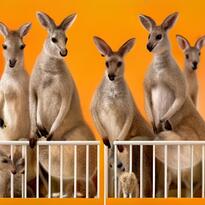}} &   
some kangaroos or some cribs, but not with both
 \\
  &  &  & 
some kangaroos or some cribs, but not with both\\
3 & \makecell{An image with some kangaroos and some cribs.\\There are as many kangaroos as cribs.}  &   &  
more kangaroos than cribs\\
 & & & 
some kangaroos or some cribs, but not with both \\
 & & &
some kangaroos or some cribs, but not with both\\
 \midrule
& & \multirow{5}{*}{\includegraphics[width=0.2\textwidth]{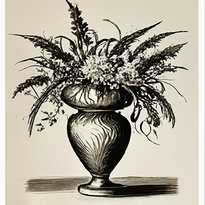}} &   
many flowers in the vase\\
  &  &  & 
 many flowers in the vase\\
4 & \makecell{An image of a vase.\\There are no flowers in the vase.}  &   &  
no flowers in the vase\\
 & & & 
only a few flowers in the vase
 \\
 & & &
only a few flowers in the vase \\
 \midrule
& & \multirow{5}{*}{\includegraphics[width=0.2\textwidth]{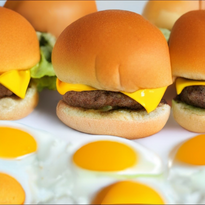}} &   
some eggs or some burgers, but not with both
 \\
  &  &  & 
some eggs or some burgers, but not with both\\
5 & \makecell{An image with some eggs and some burgers.\\There are as many eggs as burgers.}  &   &  
some eggs or some burgers, but not with both\\
 & & & 
as many eggs as burgers
 \\
 & & &
some eggs or some burgers, but not with both \\
 \midrule
& & \multirow{5}{*}{\includegraphics[width=0.2\textwidth]{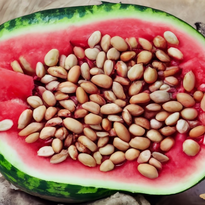}} &   
no seeds in the watermelon
 \\
  &  &  & 
many seeds in the watermelon \\
6 & \makecell{An image of a watermelon.\\ There are many seeds in the watermelon.}  &   &  
no seeds in the watermelon \\
 & & & 
no seeds in the watermelon
 \\
 & & &
no seeds in the watermelon	\\
\bottomrule
\end{tabular}}
\caption{Examples of images and annotations in Annotation Task 2. Only images, the question ``Which line describes the image the best?'' and 3--5 lines of text were shown to annotators when completing the task. The exact answers annotators selected are shortened for brevity.}
\label{fig:app_exampes_task2}
\end{figure}


%% file: tables/appendix/qual_task3.tex
\begin{figure}[h]
\renewcommand{\arraystretch}{2}
\centering
\resizebox{\textwidth}{!}{
\begin{tabular}{cccl}
\toprule
Example \# & Prompt & Image & Questions and Answers \\ \bottomrule
 & & \multirow{5}{*}{\includegraphics[width=0.2\textwidth]{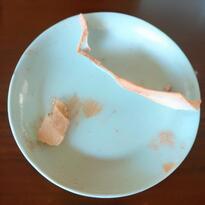}} &   
Is there a plate? 1, 1, 1, 1, 1\\
  &  &  & Is there one plate? 0, 0, 0, 1, 1 \\
1 & \makecell{There is 1 plate on the table, \\but it is broken into two pieces.}  &  &  
Are there pieces of a plate? 1, 1, 1, 0, 0 \\
 & & & 
Are there two pieces of a plate? 0, 0, 0, 0, 0\\
 & & &
Is the plate broken? 1, 1, 1, 0, 0 \\
 \midrule
  & & \multirow{5}{*}{\includegraphics[width=0.2\textwidth]{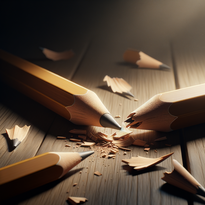}} &   
Are there pencils? 1, 1, 1, 1, 1\\
  &  &  & Are there 2 pencils? 0, 0, 0, 0, 0 \\
2 & \makecell{There are 2 pencils on the table,\\but one pencil is broken into two pieces.}  &  &  
Is one pencil broken? 0, 1, 1, 0, 0 \\
 & & & 
Is the pencil broken into 2 pieces? 0, 0, 0, 0, 0\\
 & & & \\
 \midrule
& & \multirow{5}{*}{\includegraphics[width=0.2\textwidth]{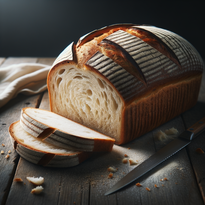}} &   
\\
  &  &  & Is there a loaf of bread? 1, 1, 1, 1, 1 \\
3 & A loaf of bread cut into thirds. &  &  
Is the loaf of bread cut into thirds? 0, 1, 0, 0, 1 \\
 & & & \\
 & & & \\  \midrule
 & & \multirow{5}{*}{\includegraphics[width=0.2\textwidth]{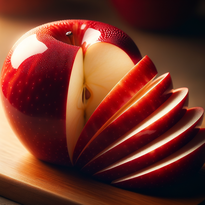}} &   
Is there an apple? 1, 1, 1, 1, 1 \\
  &  &  & Is the apple cut into pieces? 1, 1, 1, 1, 1 \\
4 & An apple cut into 5 pieces. &  &  
Are there 5 apple pieces? 1, 1, 1, 1, 1 \\
 & & & 
Are the pieces inside the apple? 0, 1, 1, 1, 0\\
 & & & \\  \midrule
 & & \multirow{5}{*}{\includegraphics[width=0.2\textwidth]{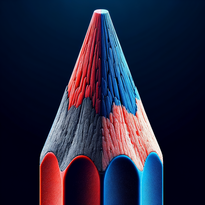}} &   
Is there a pencil? 1, 1, 1, 1, 1\\
  &  &  & Is there half of a pencil? 0, 0, 0, 0, 0 \\
5 & \makecell{An image of a pencil where one half of it is red\\and the other half is blue.}  &  &  
Is the pencil in two halves? 0, 0, 0, 0, 0 \\
 & & & 
Is one half of the pencil red? 0, 0, 1, 1, 0\\
 & & &
Is the other half of the pencil blue? 0, 0, 1, 1, 0\\    \midrule
  & & \multirow{5}{*}{\includegraphics[width=0.2\textwidth]{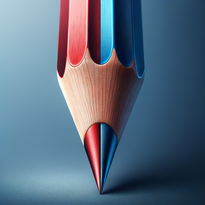}} &   
Is there a pencil? 1, 1, 1, 1, 1\\
  &  &  & Is there half of a pencil? 0, 0, 0, 0, 0 \\
6 & \makecell{An image of a pencil where one half of it is red\\and the other half is blue.}  &  &  
Is the pencil in two halves? 0, 0, 0, 0, 0 \\
 & & & 
Is one half of the pencil red? 1, 1, 1, 1, 1\\
 & & &
Is the other half of the pencil blue? 1, 1, 1, 1, 1\\  
\bottomrule
\end{tabular}}
\caption{Examples of images and annotations in Annotation Task 3. Only images and questions were shown to annotators when completing the task. The answers are encoded as numbers: ``1'' denotes ``yes'', and ``0'' denotes ``no''.}
\label{fig:app_exampes_task3}
\end{figure}

%% file: app_vqa.tex
\section{VQA experiments}
\label{app:vqa}

\subsection{The VQA Setup}

In our VQA experiments we use \paligemma~\cite{paligemma}. \paligemma is an open 3B vision-language model (VLM) inspired by PaLI-3 \cite{chen2023pali3}, built with open components, such as the SigLIP~\cite{zhai2023siglip} vision model and the Gemma-2B language model \cite{gemma2024}. We use the public checkpoint for an input resolution of $448\times 448$ pixels as counting requires fine-grained visual information available at \url{https://www.kaggle.com/models/google/paligemma/jax/paligemma-3b-pt-448}. 

To turn \geckonum into a VQA dataset/benchmark, we added questions of the form \texttt{How many <noun>s are in the image?} to each data point. As the ground truth answer we used the mode of the answers provided by the annotators, as we did in the rest of the work.

For the evaluation benchmark we only included high-quality images on which at least four out of the five annotators gave the same response; in this case, the mode corresponds to the majority vote. This yields $59,582$ high-agreement question-answer pairs from \imagen-A+B+C+D and Muse-A+B. For this set of experiments only, we also generated additional images with \imagenC, as it was one of the best performing models in this family. In total, we had 10 different images for each prompt for \imagenC. For the fine-tuning experiments we included all images even when three or more annotators disagreed in their responses. We only trained on Imagen-A+B+C+D data, which corresponds to $53,053$ ``noisy'' question-answer pairs.
In experiments where we split the dataset into ``frequent'' and ``rare'' classes, we use the filter that filters images based on the division of prompts as described in Appendix~\ref{app:prompts}.

To fine-tune \paligemma we found that fine-tuning for two epochs with a learning-rate of $10^{-5}$, a weight-decay strength of $10^{-6}$, and a batch-size of $256$ worked well across different training mixtures (otherwise we used the default hyper parameter settings). Due to the exploratory nature of these experiments, we did not re-tune hyper parameters for each experiment and setting separately, but we used a single setting for all fine-tuning experiments.

\clearpage
\subsection{Additional Results}

\subsubsection{\geckonum as a VQA benchmark}
\begin{table}[htb]
    \centering
    \caption{Accuracy (in \%) when evaluating \paligemma checkpoints on TallyQA (test) and on the \geckonum VQA benchmark.}
    \begin{tabular}{l|CC|C}
    \toprule
    \multirow{2}{*}{\paligemma checkpoint} & \multicolumn{2}{c|}{TallyQA test} & \multirow{2}{*}{\geckonum} \\
    & \text{simple} & \text{complex} & \\\midrule
    Base & 35.4 & 37.8 & 68.4 \\
    Fine-tuned (TallyQA train) & 84.9 & 72.1 & 73.3\\\bottomrule
    \end{tabular}
    \label{tab:paligemma_eval}
\end{table}
\begin{figure}[htb]
    \centering
    \begin{subfigure}[b]{0.49\textwidth}
         \centering
         \includegraphics[width=\textwidth]{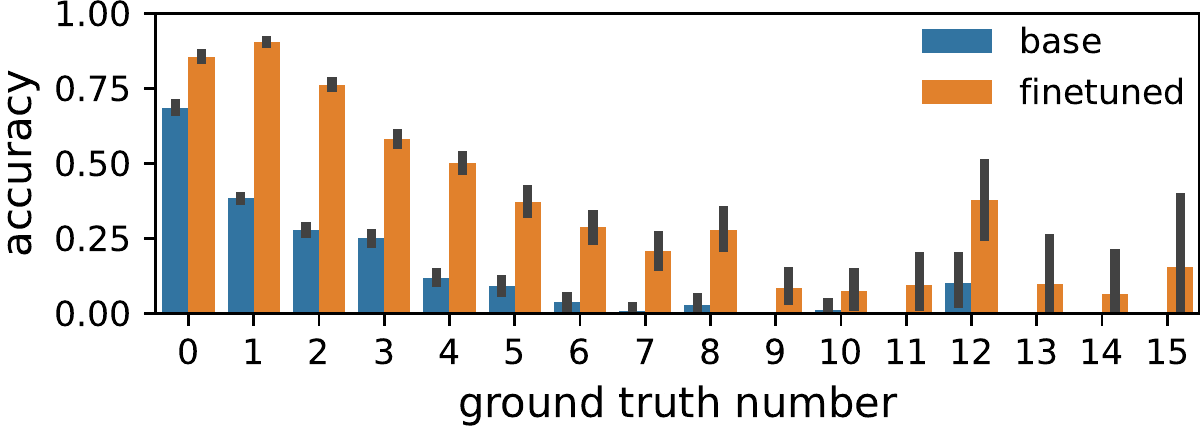}
         \caption{Accuracy on the TallyQA test set}
     \end{subfigure}
     \hfill
     \begin{subfigure}[b]{0.49\textwidth}
         \centering
         \includegraphics[width=\textwidth]{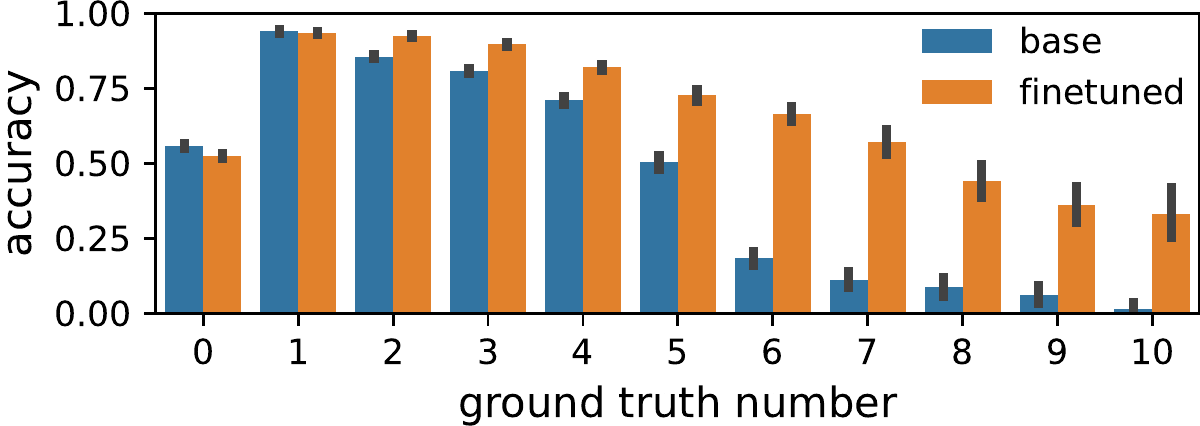}
         \caption{Accuracy on \geckonum (high agreement)}
     \end{subfigure}
    \caption{Accuracy of \paligemma (base model or fine-tuned on TallyQA (train)) when evaluated on the TallyQA test set (left) and \geckonum (right) by ground truth number.}
    \label{fig:paligemma_eval}
\end{figure}

Here we first provide additional results on using \geckonum as a VQA benchmark. We evaluate the base model and a model fine-tuned on TallyQA (train) on both TallyQA (test) as well as on \geckonum, see Table~\ref{tab:paligemma_eval} and Figure~\ref{fig:paligemma_eval} for results. 

We observe that \geckonum seems to be an easier benchmark as the base model is already able to answer many questions correctly, especially for lower counts. This is likely because while TallyQA includes complex and cluttered scenes, \geckonum images focus on one or very few object classes.

Moreover, we find that the fine-tuned model particularly improves on higher counts ($\geq 5$), as shown in Figure~\ref{fig:paligemma_eval}. We observe the same trend for both TallyQA (test) and \geckonum; this indicates that fine-tuning on TallyQA transfers to other benchmarks and datasets and that \paligemma fine-tuned has indeed improved in counting.
We note that TallyQA (test) only includes very few examples with high numbers ($\geq 9$), leading to larger uncertainties in the reported results. As we discuss in this work, all evaluated text-to-image models already struggle with much smaller numbers ($\geq 4$), and similar issues are observed with VLMs.
We speculate that as models continue to improve rapidly, benchmarks with high-quality images containing higher counts of objects will become even more important in model development and evaluation.

\subsubsection{Fine-tuning \paligemma on \geckonum}
We briefly investigate the utility of \geckonum as synthetic training data. For this we fine-tune \paligemma on different mixtures of TallyQA (train) and \imagen A+B+C+D, and evaluate these on TallyQA (test) and the split of \museB images, see Tables~\ref{tab:paligemma_finetune_exp1} and \ref{tab:paligemma_finetune_exp2}.
We have \num{53,053} \imagen-A+B+C+D images in total, which we can use for fine-tuning, while the size of TallyQA (train) is $249,318$. We perform two sets of experiments:
\begin{itemize}
    \item \textbf{Experiment 1}, in which we match \geckonum data roughly 1:1 with TallyQA data to explore the utility of the $53$k synthetic data points compared to $53$k real data points.
    \item \textbf{Experiment 2}, in which we train on all available TallyQA data and add \geckonum data to explore the benefits of enlarging an established training set with synthetic data.
\end{itemize}

In \textbf{Experiment 1} we control for fine-tuning set sizes and pair the $53$k \imagen images with $53$k randomly sampled TallyQA train images (for a total of $106$k examples) and compare them to fine-tuning on only $106$k randomly sampled TallyQA train images. We also consider paring them with half the amount of TallyQA images ($26$k) as well as only training on the \imagen images. To control for the fact that \imagen and Muse share the same object classes, we also consider a split of \imagen and Muse into two disjoint object sets of ``frequent'' and ``rare'' as described above and only train on the ``frequent'' subset and treating the ``rare'' subset as a held-out set. 

The results in Table~\ref{tab:paligemma_finetune_exp1} can be summarized as:
\begin{itemize}
    \item Including \geckonum \imagen images \emph{in addition to} TallyQA images neither boosts nor deteriorates performance on TallyQA (test) significantly; there is a slight improvement on TallyQA (test) ``simple'' and a slight deterioration on TallyQA (test) ``complex''. 
    \item However, \emph{replacing} some or all of the TallyQA (train) images with \imagen images does deteriorate performance on TallyQA (test) by up to a few percentage points.
    \item Including \imagen images always  markedly improves performance on \museB (by more than $25$ \pp). This is still true when only training on ``frequent'' classes while evaluating only on ``rare'' classes, though the improvement is slightly smaller in this case (by about $20$ \pp). There is clear generalization from frequent to rare classes.
    \item Fine-tuning on \imagen only already leads to strong performance on TallyQA (test) ``simple'' (about $70\%$ accuracy compared to $35\%$ for the base model and $82\%$ when training on TallyQA) but not ``complex'' (only an increase to $44.9\%$ from $37.8\%$ compared to $67.8\%$ when training on TallyQA). The latter makes intuitive sense as the \geckonum VQA questions resemble ``simple'' TallyQA questions (\texttt{How many <object> are there in the image?}) but not ``complex'' ones (\texttt{How many <object>s have <property>?}, e.g. ``How many giraffes are lying down?'').
\end{itemize}
Thus, as noted in Section~\ref{sec:challenges}, adding \imagen images to the training mix does not hurt performance on TallyQA (test), but it allows us to achieve much better results on other data splits not covered by the TallyQA training data as well (the \museB split in this case).
We also found that during fine-tuning on TallyQA (train) the model slightly overfits to the TallyQA dataset in that the best TallyQA (test) performance is achieved at the end of fine-tuning while the best performance on \museB is achieved partway through fine-tuning and performance reduces slightly as fine-tuning continues.

In \textbf{Experiment 2}, we use the full TallyQA (train) set and add the $53$k \imagen images in addition, and compare this mixture to slight variations where we account for the number of training examples (by removing $53$k examples from TallyQA) and the number of training steps (by training on TallyQA for longer).
The results are shown in Table~\ref{tab:paligemma_finetune_exp2}.
We replicate previous results and observe the same trends; the changes on TallyQA (test) performance are very small ($<1\%$ in the worst case but usually much smaller) when including \imagen images while performance on \museB is drastically improved even in the held-out case.

\input{./tables/appendix/paligemma_acc}
\clearpage


%% file: tables/appendix/paligemma_acc.tex
\begin{table}[htb]
    \centering
    \caption{Accuracy of \paligemma fine-tuned on different data mixtures (at resolution $448\times 448$ pixels). Here, we only train on a subset of TallyQA (train), which in total has approximately 250k examples. \imagen here means \imagen A+B+C+D. We provide the approximate number of unique data points in brackets.}\label{tab:paligemma_finetune_exp1}
    \begin{tabular}{l|CC|CCC}
    \toprule
    \multirow{2}{*}{Fine-tuning data} & \multicolumn{2}{c|}{TallyQA test} & \multicolumn{3}{c}{\museB} \\
    & \text{simple} & \text{complex} & \text{all} & \text{\,freq.\,} & \text{rare} \\\midrule
    TallyQA (106k) &83.1 & 69.4 & 70.3 & 76.5 & 63.2 \\
    TallyQA (53k) & 82.0 & 67.8 & 68.3 & 75.0 & 60.5 \\
    TallyQA (53k) + \imagen (freq \& rare, 53k) & 82.4 & 67.8 & 89.4 & 90.0 & 88.8  \\
    TallyQA (53k) + \imagen (only freq, 28k) & 82.2 & 68.3 & 86.0 & 89.5 & 82.0 \\
    TallyQA (26k) & 81.1 & 66.5 & 68.7 & 75.2 & 61.3 \\
    TallyQA (26k) + \imagen (freq \& rare, 53k) & 81.5 & 66.4 & 89.6 & 90.0 & 89.2 \\
    \imagen (freq \& rare, 53k) & 70.7 & 44.9 & 90.0 & 90.1 & 89.7 \\
    \imagen (only freq, 28k) & 70.1 & 44.7 & 85.0 & 88.6 & 80.7 \\
    \midrule
    Base model (no fine-tuning) &35.4 & 37.8&67.0&71.0&62.4 \\\bottomrule
    \end{tabular}
    
\end{table}

\begin{table}[htb]
    \centering
    \caption{Accuracy (in $\%$) of \paligemma fine-tuned on different data mixtures (at resolution $448\times 448$ pixels). Here, we train on all TallyQA (train) images. \imagen here means \imagen-A+B+C+D and ``all'' refers to freq. and rareuent objects.}\label{tab:paligemma_finetune_exp2}    
    \begin{tabular}{lc|CC|CCC}
    \toprule
    \multirow{2}{*}{Fine-tuning data} & \text{data} & \multicolumn{2}{c|}{TallyQA test} & \multicolumn{3}{c}{\museB} \\
    & \text{points} & \text{simple} & \text{complex} & \text{all} & \text{only freq.} & \text{rare} \\\midrule
    TallyQA &250k & 84.7 & 72.8 & 71.1 & 76.6 & 64.8\\
    TallyQA (only 200k) + \imagen (all) & 250k & 84.7 & 71.9 & 90.6 & 90.9 & 90.3 \\
    TallyQA + \imagen (all) & 300k& 84.9 & 72.7 & 90.5 & 91.2 & 89.8 \\
    TallyQA + \imagen (only freq) & 275k& 84.9 & 72.5 & 87.3 & 90.4& 83.6 \\
    TallyQA (longer training) & 250k & 84.7 & 72.4 & 69.8 & 76.9 & 61.7 \\\midrule
    Base model (no fine-tuning) & --&35.4 & 37.8&67.0&71.0&62.4 \\\bottomrule
    \end{tabular}
\end{table}